\documentclass[10pt,conference]{IEEEtran}
\IEEEoverridecommandlockouts

\usepackage{cite}
\usepackage{amsmath,amssymb,amsfonts}
\usepackage{algorithmic}
\usepackage{graphicx}
\usepackage{textcomp}

\usepackage{amsmath}
\usepackage{algorithm}
\usepackage{algorithmic}
\usepackage{array}
\usepackage[dvipsnames]{xcolor}
\usepackage[hidelinks]{hyperref}
\usepackage{booktabs}

\def\BibTeX{{\rm B\kern-.05em{\sc i\kern-.025em b}\kern-.08em
    T\kern-.1667em\lower.7ex\hbox{E}\kern-.125emX}}

\begin{document}

\title{RO-FIGS: Efficient and Expressive Tree-Based Ensembles for Tabular Data
\thanks{\textbf{© 2025 IEEE. Personal use of this material is permitted. Permission from IEEE must be obtained for all other uses, in any current or future media, including reprinting/republishing this material for advertising or promotional purposes, creating new collective works, for resale or redistribution to servers or lists, or reuse of any copyrighted component of this work in other works. }\newline \newline
UM is grateful to the Cambridge Commonwealth, European \& International Trust for funding received as a DeepMind Cambridge Scholar to undertake her degree, and for the Trust’s and Google DeepMind’s support in publishing this article; UM also acknowledges the support from the Department of Computer Science and Technology, University of Cambridge, UK. NS and MJ acknowledge the support of the U.S. Army Medical Research and Development Command of the Department of Defense; through the FY22 Breast Cancer Research Program of the Congressionally Directed Medical Research Programs, Clinical Research Extension Award GRANT13769713. Opinions, interpretations, conclusions, and recommendations are those of the authors and are not necessarily endorsed by the Department of Defense.}
}

\author{
\IEEEauthorblockN{Ur\v{s}ka Matja\v{s}ec$^1$\quad\quad\quad\quad\quad Nikola Simidjievski$^{2,1}$\quad\quad\quad\quad\quad Mateja Jamnik$^1$}
\IEEEauthorblockA{$^1$\textit{Department of Computer Science and Technology}, $^2$PBCI, \textit{Department of Oncology}\\
\textit{University of Cambridge, Cambridge, UK}\\
\{um234,ns779,mj201\}@cam.ac.uk}
}

\maketitle

\begin{abstract}
Tree-based models are often robust to uninformative features and can accurately capture non-smooth, complex decision boundaries. Consequently, they often outperform neural network-based models on tabular datasets at a significantly lower computational cost. Nevertheless, the capability of traditional tree-based ensembles to express complex relationships efficiently is limited by using a single feature to make splits. To improve the efficiency and expressiveness of tree-based methods, we propose Random Oblique Fast Interpretable Greedy-Tree Sums (\mbox{RO-FIGS}). \mbox{RO-FIGS} builds on Fast Interpretable Greedy-Tree Sums, and extends it by learning trees with oblique or multivariate splits, where each split consists of a linear combination learnt from random subsets of features. This helps uncover interactions between features and improves performance. The proposed method is suitable for tabular datasets with both numerical and categorical features. We evaluate \mbox{RO-FIGS} on 22 real-world tabular datasets, demonstrating superior performance and much smaller models over other tree- and neural network-based methods. Additionally, we analyse their splits to reveal valuable insights into feature interactions, enriching the information learnt from SHAP summary plots, and thereby demonstrating the enhanced interpretability of \mbox{RO-FIGS} models. The proposed method is well-suited for applications, where balance between accuracy and interpretability is essential.
\end{abstract}


\section{Introduction}
\label{sec:introduction}

Current state-of-the-art methods for tabular data include gradient-boosted decision trees (GBDTs)~\cite{xgboost, ke_2017_lightgbm, catboost} and several neural network-based methods such as FT-Transformer~\cite{gorishniy_2021_revisiting_dl_models} and TabPFN~\cite{hollmann_2023_tabpfn}. However, no single method consistently outperforms the others~\cite{shwartz_2021_dl_not_all_you_need,borisov_2022_dnn_and_tabular_survey,grinsztajn_2022_trees_outperform_dl,mcelfresh_2023_tabzilla}. We focus on improving tree-based methods since they often have a computational advantage over complex neural networks with millions of parameters. Additionally, their hierarchical structure makes tree-based methods easier to comprehend than deep neural networks. Nevertheless, boosted ensembles, which iteratively add new trees to the model to correct the errors of previously built trees~\cite{xgboost, ke_2017_lightgbm, catboost}, can become complex with hundreds of trees and splits, making their interpretation challenging. Moreover, traditional splitting methods in boosted ensembles rely on single features, limiting the model's ability to capture complex relationships between features.\looseness-1

To overcome these limitations, we introduce \mbox{RO-FIGS}: a random oblique fast interpretable greedy-tree sum method, which constructs splits as linear combinations of features, rather than considering only a singular feature to make a split. The splits reflect interactions among multiple features, which makes them more expressive than univariate splits. Moreover, RO-FIGS captures complex relationships between features more efficiently, as fewer splits are needed to represent complex patterns, leading to more compact models.

The structure of \mbox{RO-FIGS} builds on FIGS~\cite{tan_2022_figs}, which allows for constructing models in a flexible way as it can adapt to the additive structure if one exists in the data. However, it differs from FIGS due to the following technical novelties. First, \mbox{RO-FIGS} employs oblique splits, thereby enabling feature interaction directly within the splits. Second, it adopts a different stopping condition. Instead of strictly limiting the number of splits, \mbox{RO-FIGS} uses the minimum impurity decrease value to stop the training, which potentially yields larger models but also significantly improves the performance. Nevertheless, the size of both FIGS and \mbox{RO-FIGS} models remains comparable across most datasets.

We demonstrate the compactness of \mbox{RO-FIGS} models, consisting of up to only five trees and, with few exceptions, a low number of splits. This compact nature of \mbox{RO-FIGS} models not only improves their efficiency, which is measured as the balance between performance and compactness, but also makes them inherently more interpretable. Furthermore, by analysing linear combinations of features that often appear together in oblique splits, we offer a better understanding of the importance and interaction between features. We show that these splits are expressive and further enhance the interpretability of \mbox{RO-FIGS}. This suggests that \mbox{RO-FIGS} is well-suited for real-world scenarios, offering a good balance between performance and interpretability. 

The contributions of this paper are two-fold:
\begin{enumerate}
    \item We introduce \mbox{RO-FIGS}, a novel and well-performing tree-based method, which builds an ensemble of oblique trees in an additive way (Section~\ref{sec:method}). The implementation is available at: {\color{RoyalBlue}\url{https://github.com/um-k/rofigs}}.
    \item We demonstrate the capabilities of RO-FIGS in a series of experiments, showing they produce compact and performant models at lower computational cost, compared to many state-of-the-art methods for tabular data (Sections \ref{sec:accuracy} and \ref{sec:efficiency}). Further analysis reveals \mbox{RO-FIGS'} interpretability properties beyond those produced by standard post-hoc approaches (such as SHAP), thus further enhancing our understanding of the model's decision-making process (Section \ref{sec:expressiveness}). \looseness -1

\end{enumerate}

\section{Related work}
\label{sec:related-work}

\paragraph*{Greedy-tree sums} The foundation of our proposed method are greedy-tree sums (FIGS)~\cite{tan_2022_figs}, a generalisation of CART algorithm~\cite{cart_1984}, which allows for building multiple trees in an additive way. This process involves creating new trees using the residuals of the preceding ones, much like GBDTs. However, FIGS adds one split to the model in each iteration, in contrast to GBDTs, which add the entire trees. There are two stopping criteria for FIGS: the number of trees and the total number of splits across all trees. FIGS outperforms traditional decision tree models, while at the same time building smaller, more compact models. 
Tan et al.~\cite{tan_2022_figs} show that FIGS outperforms boosted stumps, but our experiments show that its performance is still subpar to traditional GBDTs like LightGBM~\cite{ke_2017_lightgbm} and CatBoost~\cite{catboost} (see Table~\ref{tab__performance}). The strict stopping conditions prevent FIGS from undergoing sufficient training, leading to its underperformance. Instead of applying strict constraints on the number of trees and splits in the model, \mbox{RO-FIGS} is more flexible and employs a minimum impurity decrease value to stop the training process. 

\paragraph*{Oblique trees} While traditional decision trees use a single feature in any splitting node, oblique or multivariate trees use a combination of features to define a split~\cite{cart_1984,murthy_1994_oc1,yang_2019_wodt}. This makes them more expressive than univariate trees and enables them to learn similar decision boundaries with much shallower trees. However, oblique splits are more complex and can be harder to comprehend than univariate splits. This is especially pronounced when a split represents a linear combination of a substantial number of features. To reduce the complexity of splits, we use a hyperparameter to limit the number of features per split and employ further regularisation inside the splits.

\paragraph*{Ensembles} Tree-based models hierarchically split the data based on feature values. Tree ensembles often achieve better performance and robustness by learning and combining many trees, either simultaneously (e.g., random forest~\cite{breiman_2001}) or sequentially (e.g., FIGS~\cite{tan_2022_figs}, LightGBM~\cite{ke_2017_lightgbm}). \mbox{RO-FIGS} builds on sequential approaches, where new ensemble-constituent trees are learnt using the errors made by preceding trees, using their residuals. Ensembles of decision trees, both traditional and oblique, generally outperform single trees at the cost of increased model complexity and reduced interpretability. We avoid this problem by using regularisation within the splits. 

\paragraph*{Optimal trees} Unlike greedy algorithms that employ heuristics for tree construction, optimal tree methods~\cite{ot_oct, ot_dl8, ot_binoct, ot_gosdt} seek to learn the globally optimal trees, which can result in superior performance. However, their computational demand can often hinder their applicability to larger datasets. This limitation arises from the exhaustive search required to identify the best possible tree configuration. To mitigate this, some optimal tree methods~\cite{ot_narodytska, ot_avellaneda, ot_firat} may introduce constraints that compromise optimality in favour of computational efficiency. While optimal trees are generally small in size, we show that \mbox{RO-FIGS} has a smaller trade-off between performance and both size and computational cost.

\section{Method}
\label{sec:method}

We denote a training set as $\mathit{D := \{(\textbf{x}_i, y_i)\}_{i=1}^n}$, where $n$ is the number of samples and $y_i$ are the corresponding labels that can be continuous values in regression tasks or binary labels (0 or 1) in binary classification tasks. 

\begin{figure}[t]   
    \centering
    \includegraphics[width=0.5\textwidth]{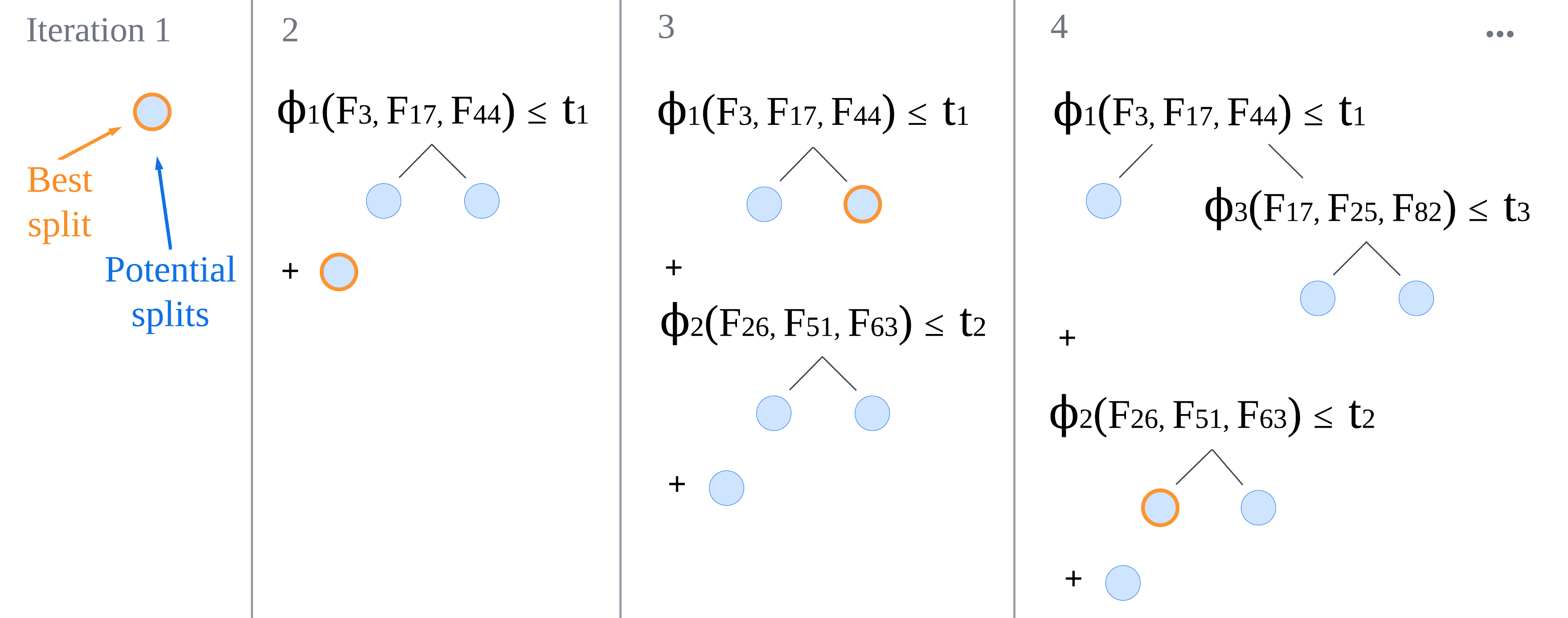}
    \caption{Fitting process of \mbox{RO-FIGS} on a toy example. \mbox{RO-FIGS} adds one split to the model in each iteration. Splits are oblique and computed as linear combinations $\Phi()$  of multiple randomly selected features $F$. $t$ denotes the threshold value for splitting. The figure has been adapted from~\cite{tan_2022_figs}.}
    \label{fig__rofigs_only} 
\end{figure}

\mbox{RO-FIGS} leverages the concept of FIGS and constructs a flexible number of (usually a few) decision trees in summation. This means that at inference time, predictions are made by summing the predictions of all trees, while additionally applying a sigmoid function for classification tasks. 
The pseudocode of \mbox{RO-FIGS} is shown in Algorithm~\ref{alg__rofigs}, and Fig.~\ref{fig__rofigs_only} illustrates the learning procedure in \mbox{RO-FIGS}.

\begin{algorithm}[tb!]
    \caption{RO-FIGS}
    \label{alg__rofigs}
\small{

\begin{algorithmic}[1]
    \STATE {\bfseries Input:} X: features, y: outcomes, \textsc{beam\_size}: number of considered features per split, \textsc{min\_imp\_dec}: minimum impurity decrease, \textsc{max\_splits}: maximum number of splits

    \STATE $\mathit{trees}$ = []

    \WHILE {(get\_max\_imp\_dec() $>$ \textsc{min\_imp\_dec} \OR first\_iteration) \AND count\_total\_splits() $<$ \textsc{max\_splits}}

        \STATE $\mathit{feat}$ = select\_random(\textsc{beam\_size})
        \STATE $\Phi$ = compute\_linear\_combination($\mathit{X, y, feat}$)
        \STATE $\mathit{all\_trees}$ = join($\mathit{trees}$, define\_oblique\_split($\Phi$))
        
        \STATE $\mathit{potential\_splits}$ = []
        \FOR{$\mathit{tree}$ in $\mathit{all\_trees}$}
            \STATE $\mathit{y\_res}$ = $\mathit{y}$ – predict($\mathit{all\_trees}$ except $\mathit{tree}$) \hspace{0.55cm}// residuals

            \FOR{repetition in range($r$)}
                \STATE $\mathit{feat}$ = select\_random(\textsc{beam\_size})
                \FOR{$\mathit{leaf}$ in $\mathit{tree}$}
                  
                    \STATE $\Phi$ = compute\_linear\_combination($\mathit{X, y\_res, feat}$)
                    \STATE $\mathit{potential\_split}$ = define\_oblique\_split($\Phi, \mathit{leaf}$)          
                    \STATE $\mathit{potential\_splits}$.append($\mathit{potential\_split}$) 
                \ENDFOR

                \STATE $\mathit{best\_split}$ = split\_with\_max\_imp\_dec($\mathit{potential\_splits}$)

                \IF{impurity\_decrease($\mathit{best\_split}$) $ > $ \textsc{min\_imp\_dec}}
                    \STATE break
                \ENDIF
            
            \ENDFOR
        \ENDFOR
        \STATE $\mathit{trees}$.insert($\mathit{best\_split}$)
   \ENDWHILE

   \STATE {\bfseries Return:} $\mathit{trees}$
\end{algorithmic}
}
\end{algorithm}

In each iteration, \mbox{RO-FIGS} adds one split to the existing model (line 23). This split either replaces one of the current leaves or adds an additional stump to the model. 
At the start of every iteration, \mbox{RO-FIGS} randomly selects a subset of features that will be considered for learning splits in this iteration (line~11). The size of the subsets, which we refer to as {\em beams}, is determined by hyperparameter \textsc{beam\_size}, which can take any value between one and the number of all available features. 
At iteration $i$, there are $i+j$ potential splits, where $j$ refers to the current number of trees in the model (lines~\mbox{13-15}). A potential split with the maximum impurity decrease is then added to the current model (lines 17 and 23). 
Due to the random feature selection, it is possible that only uninformative features are chosen. We, therefore, allow multiple repetitions per iteration. More specifically, if all $i+j$ potential splits have impurity decrease lower than the minimum impurity decrease value (i.e., \textsc{min\_imp\_dec}; line~18), then another subset of features is randomly selected and evaluated. The process is repeated up to $r$ times per iteration (line 10).

\subsection{Split optimisation} 
Each potential split is an oblique stump computed on a random subset of features. A linear combination of $k$ features is represented as $\Phi(\mathit{F_1, \dots, F_k})$ (line 13, $\mathit{k = \textsc{beam\_size}}$). We use the implementation of Stepi\v{s}nik and Kocev~\cite{stepisnik_2021_oblique_pcts} (with the default hyperparameters, except for choosing one tree with depth one) to create splits with \mbox{gradient-descent-based} split learning. The computed splits take form of $\mathit{w_1*F_1 + \dots + w_k*F_k \leq t}$, where $w_1$ and $w_k$ represent weights, $F_1$ and $F_k$ features, and $t$ a threshold. The concrete steps of the split optimisation are as follows. First, the splitting features are randomly chosen (i.e., $\mathit{F_1, \dots, F_k}$). The objective function for the split optimisation is defined as\looseness=-1
\begin{equation}
\label{eq:split_optimisation}
    min_{w, b} ||w||_{1/2} + C*g(w, b),
\end{equation}
where $||w||_{1/2} = (\sum_{i=1}^{k} \sqrt{|w_i|})^2$, $k$ is the number of features, and $C$ corresponds to the regularisation strength. L$\frac{1}{2}$ norm induces weight sparsity and reduces split complexity, resulting in smaller splits. 
A fitness function $g$ guides the optimisation process with the objective to minimise the impurity (defined as weighted variances) on both sides of the hyperplane.

\subsection{Nominal features}
We explore three encoding techniques for transforming nominal features (i.e., categorical features without an inherent order) into numerical ones. The first approach, $\textsc{E-ohe}$, employs one-hot encoding, a common practice for handling nominal features~\cite{scikit_learn}. The other two techniques, denoted as $\textsc{E-count}$ and $\textsc{E-prop}$, are target encoding variants that leverage target information to transform nominal features into ordered ones in such an order that the target consistently changes~\cite{target_encoding_barreca, target_encoding_wright, targeto_encoding_zhu}. Specifically, $\textsc{E-prop}$ orders categories based on the proportion of 1s in binary classification tasks (or by increasing mean in regression tasks), while $\textsc{E-count}$ simply counts 1s in the target variable. The main advantage of $\textsc{E-count}$ and $\textsc{E-prop}$ over $\textsc{E-ohe}$ is that they do not increase the dimensionality of the data. However, $\textsc{E-ohe}$ leads to best performance on average (see Table~\ref{tab__encodings} in Appendix~\ref{sec:appendix-experiments:datasets}\footnote{Supplementary material containing all appendices is available at: {\mbox{\color{RoyalBlue}\url{https://github.com/um-k/rofigs}}}.}), so we one-hot encode nominal features before passing them to \mbox{RO-FIGS}.

\subsection{Hyperparameters and stopping conditions} \mbox{RO-FIGS}’s primary stopping condition is based on the minimum impurity decrease (i.e., \textsc{min\_imp\_dec}), a standard parameter in tree-based methods. More specifically, in the case of RO-FIGS, this is further regularised, where the training process stops if there is no significant impurity decrease (specified value) after $r$ repeated attempts (five by default). Note that we did not observe any performance benefits when employing other criteria here, such as limiting the number of splits as in~\cite{tan_2022_figs} (see Table~\ref{tab__stopping_conditions} in Appendix~\ref{sec:appendix-results:ablations}). \mbox{RO-FIGS} also employs a hyperparameter for \textsc{beam\_size}, which determines the size of the beam, that is, the number of features considered per split. This parameter influences the complexity of individual splits: larger values allow for more exploration but increase computational cost, while smaller values may restrict optimal splits. \looseness -1

\section{Experiments}
\label{sec:experiments}

We quantitatively evaluate \mbox{RO-FIGS} (Section~\ref{sec:accuracy}) and discuss the efficiency of tree-based models by analysing the balance between their accuracy and size (Section~\ref{sec:efficiency}). We inspect oblique splits to show the expressiveness and enhanced interpretability of RO-FIGS models (Section~\ref{sec:expressiveness}). 

\begin{table*}[th!]\setlength{\tabcolsep}{2.5pt}
    \caption{Classification performance of RO-FIGS and 10 baselines on 22 tabular datasets. We report mean $\pm$ std of the test balanced accuracy across 10 folds, and average rank across all datasets. Note that we use reversed rank (where higher is better). We highlight the \textcolor{ForestGreen}{\textbf{first}}, \textcolor{Bittersweet}{\textbf{second}}, and \textcolor{Mulberry}{\textbf{third}} highest accuracy for each dataset. RO-FIGS is amongst the top three on 11 datasets and ranks the best overall.}
    \centering
    \resizebox{\textwidth}{!}{
    \begin{tabular}{lrrrrrrrrrrr}
        \toprule
 
        Dataset & DT & MT & OT & ODT & RF & ETC & CatBoost & FIGS & Ens-ODT & MLP & \textbf{RO-FIGS} \\
        \midrule
        blood & 63.3 $_{\pm  7.3}$ & \textcolor{Mulberry}{\textbf{64.5 $_{\pm  2.6}$}} & 64.1 $_{\pm  3.6}$ & 61.5 $_{\pm  8.9}$ & 63.0 $_{\pm  5.6}$ & 58.0 $_{\pm  5.6}$ & 60.3 $_{\pm  5.8}$ & \textcolor{Bittersweet}{\textbf{65.0 $_{\pm  4.9}$}} & 62.8 $_{\pm  4.4}$ & 55.8 $_{\pm  5.1}$ & \textcolor{ForestGreen}{\textbf{68.5 $_{\pm  4.1}$}} \\

        diabetes & 70.8 $_{\pm  4.9}$ & 68.5 $_{\pm  5.8}$ & 71.0 $_{\pm  7.1}$ & 69.7 $_{\pm  8.3}$ & \textcolor{Bittersweet}{\textbf{72.6 $_{\pm  5.7}$}} & \textcolor{Mulberry}{\textbf{72.5 $_{\pm  6.2}$}} & 70.5 $_{\pm  7.0}$ & 70.2 $_{\pm  4.8}$ & 69.9 $_{\pm  4.8}$ & 70.6 $_{\pm  6.5}$ & \textcolor{ForestGreen}{\textbf{73.6 $_{\pm  4.7}$}} \\
        
        breast-w & 93.5 $_{\pm  3.0}$ & \textcolor{ForestGreen}{\textbf{96.5 $_{\pm  2.0}$}} & 93.9 $_{\pm  2.2}$ & 93.4 $_{\pm  4.0}$ & 95.8 $_{\pm  1.8}$ & \textcolor{ForestGreen}{\textbf{96.5 $_{\pm  1.6}$}} & 95.9 $_{\pm  2.8}$ & 94.4 $_{\pm  2.4}$ & \textcolor{Bittersweet}{\textbf{96.2 $_{\pm  1.5}$}} & \textcolor{Mulberry}{\textbf{96.1 $_{\pm  1.5}$}} & \textcolor{ForestGreen}{\textbf{96.5 $_{\pm  1.9}$}} \\
        
        ilpd & 56.0 $_{\pm  4.1}$ & 57.3 $_{\pm  5.5}$ & 52.0 $_{\pm  3.8}$ & 54.2 $_{\pm  10.0}$ & 56.6 $_{\pm  6.9}$ & 54.6 $_{\pm  4.9}$ & \textcolor{Mulberry}{\textbf{58.9 $_{\pm  6.8}$}} & 53.2 $_{\pm  7.3}$ & \textcolor{Bittersweet}{\textbf{59.8 $_{\pm  4.9}$}} & 51.6 $_{\pm  3.7}$ & \textcolor{ForestGreen}{\textbf{61.9 $_{\pm  10.4}$}} \\
        
        monks2 & 94.6 $_{\pm  4.8}$ & 87.1 $_{\pm  8.9}$ & 52.8 $_{\pm  5.4}$ & 86.8 $_{\pm  14.6}$ & 74.9 $_{\pm  6.3}$ & 91.1 $_{\pm  5.3}$ & 78.3 $_{\pm  8.1}$ & 63.7 $_{\pm  7.6}$ & \textcolor{Mulberry}{\textbf{98.1 $_{\pm  2.5}$}} & \textcolor{ForestGreen}{\textbf{99.9 $_{\pm  0.1}$}} & \textcolor{Bittersweet}{\textbf{99.5 $_{\pm  0.8}$}} \\
        
        climate & 69.4 $_{\pm  11.6}$ & 69.8 $_{\pm  6.3}$ & 62.9 $_{\pm  10.9}$ & \textcolor{Mulberry}{\textbf{71.4 $_{\pm  11.1}$}} & 51.0 $_{\pm  3.2}$ & 50.0 $_{\pm  0.0}$ & 68.1 $_{\pm  10.7}$ & 64.8 $_{\pm  10.8}$ & 71.2 $_{\pm  12.2}$ & \textcolor{ForestGreen}{\textbf{76.8 $_{\pm  12.5}$}} &  \textcolor{Bittersweet}{\textbf{73.1 $_{\pm  15.6}$}} \\
        
        kc2 & \textcolor{Mulberry}{\textbf{72.4 $_{\pm  10.0}$}} & 66.6 $_{\pm  6.8}$ & 70.2 $_{\pm  6.8}$ & 68.6 $_{\pm  9.2}$ & 72.1 $_{\pm  6.0}$ & 67.1 $_{\pm  9.0}$ & 69.2 $_{\pm  7.5}$ & \textcolor{Bittersweet}{\textbf{72.8 $_{\pm  10.5}$}} & 69.6 $_{\pm  6.2}$ & 69.6 $_{\pm  7.3}$ & \textcolor{ForestGreen}{\textbf{77.6 $_{\pm  7.9}$}} \\
        
        pc1 & \textcolor{Mulberry}{\textbf{60.4 $_{\pm  4.9}$}} & 53.1 $_{\pm  4.7}$ & 56.6 $_{\pm  7.4}$ & 58.5 $_{\pm  7.5}$ & 56.9 $_{\pm  5.7}$ & 56.4 $_{\pm  6.8}$ & \textcolor{Bittersweet}{\textbf{61.2 $_{\pm  9.4}$}} & 56.4 $_{\pm  5.5}$ & 60.0 $_{\pm  5.3}$ & 50.5 $_{\pm  1.7}$ & \textcolor{ForestGreen}{\textbf{66.6 $_{\pm  7.2}$}} \\
        
        kc1 & \textcolor{Mulberry}{\textbf{60.8 $_{\pm  3.3}$}} & 58.7 $_{\pm  4.2}$ & 58.7 $_{\pm  3.9}$ & 55.8 $_{\pm  5.8}$ & 60.5 $_{\pm  4.7}$ & 59.2 $_{\pm  5.6}$ & \textcolor{ForestGreen}{\textbf{65.0 $_{\pm  3.3}$}} & 59.0 $_{\pm  4.9}$ & 60.3 $_{\pm  4.2}$ & 57.9 $_{\pm  3.5}$ & \textcolor{Bittersweet}{\textbf{64.9 $_{\pm  4.9}$}} \\
        
        heart & 80.0 $_{\pm  6.9}$ & 78.2 $_{\pm  6.2}$ & 82.5 $_{\pm  4.4}$ & 77.8 $_{\pm  5.8}$ & 82.5 $_{\pm  7.5}$ & 82.7 $_{\pm  7.5}$ & \textcolor{Mulberry}{\textbf{83.3 $_{\pm  5.2}$}} &  81.2 $_{\pm  7.6}$ & \textcolor{Bittersweet}{\textbf{84.2 $_{\pm  8.4}$}} &  82.5 $_{\pm  6.4}$ & \textcolor{ForestGreen}{\textbf{84.8 $_{\pm  6.4}$}} \\
        
        tictactoe & 92.2 $_{\pm  2.5}$ & \textcolor{Bittersweet}{\textbf{97.6 $_{\pm  1.8}$}} &  67.6 $_{\pm  3.7}$ & 86.7 $_{\pm  7.0}$ & 95.9 $_{\pm  4.3}$ & 97.4 $_{\pm  3.2}$ & \textcolor{ForestGreen}{\textbf{99.9 $_{\pm  0.1}$}} &  84.3 $_{\pm  5.7}$ & 97.1 $_{\pm  2.5}$ & \textcolor{Mulberry}{\textbf{97.5 $_{\pm  1.7}$}} &  94.4 $_{\pm  3.1 }$ \\
        
        wdbc & 91.8 $_{\pm  4.3}$ & 95.2 $_{\pm  3.8}$ & 92.0 $_{\pm  4.2}$ & 95.4 $_{\pm  3.8}$ & 94.6 $_{\pm  3.5}$ & 94.3 $_{\pm  3.2}$ & \textcolor{Bittersweet}{\textbf{96.3 $_{\pm  2.4}$}} &  92.6 $_{\pm  3.5}$ & \textcolor{Mulberry}{\textbf{95.8 $_{\pm  3.7}$}} &  \textcolor{ForestGreen}{\textbf{97.2 $_{\pm  3.1}$}} &  95.6 $_{\pm  3.5 }$ \\
        
        churn & 84.2 $_{\pm  1.8}$ & 82.9 $_{\pm  2.1}$ & 67.5 $_{\pm  1.7}$ & \textcolor{Mulberry}{\textbf{86.7 $_{\pm  1.8}$}} &  83.8 $_{\pm  2.1}$ & 74.4 $_{\pm  3.2}$ & \textcolor{ForestGreen}{\textbf{88.3 $_{\pm  1.9}$}} &  82.6 $_{\pm  3.6}$ & \textcolor{Bittersweet}{\textbf{87.8 $_{\pm  2.4}$}} &  84.8 $_{\pm  3.6}$ & 86.3 $_{\pm  2.0 }$ \\
        
        pc3 & 58.3 $_{\pm  4.4}$ & \textcolor{Bittersweet}{\textbf{59.7 $_{\pm  6.1}$}} &  52.3 $_{\pm  2.7}$ & 56.8 $_{\pm  6.3}$ & 52.3 $_{\pm  3.3}$ & 52.0 $_{\pm  2.4}$ & \textcolor{Bittersweet}{\textbf{60.7 $_{\pm  6.5}$}} &  54.4 $_{\pm  4.6}$ & 55.6 $_{\pm  4.1}$ & 50.0 $_{\pm  0.0}$ & \textcolor{ForestGreen}{\textbf{62.9 $_{\pm  6.6}$}} \\

        biodeg & 78.3 $_{\pm  4.0}$ & 82.1 $_{\pm  3.0}$ & 76.5 $_{\pm  4.5}$ & 79.4 $_{\pm  2.7}$ & 82.0 $_{\pm  3.1}$ & 81.4 $_{\pm  4.9}$ & \textcolor{Bittersweet}{\textbf{85.3 $_{\pm  2.7}$}} &  76.6 $_{\pm  5.1}$ & \textcolor{Mulberry}{\textbf{83.9 $_{\pm  4.2}$}} &  \textcolor{ForestGreen}{\textbf{86.2 $_{\pm  5.6}$}} & 82.7 $_{\pm  3.7 }$ \\
        
        credit & 83.9 $_{\pm  3.8}$ & 82.0 $_{\pm  3.9}$ & 85.4 $_{\pm  4.0}$ & 84.3 $_{\pm  3.7}$ & \textcolor{Bittersweet}{\textbf{86.4 $_{\pm  5.2}$}} &  \textcolor{Mulberry}{\textbf{86.1 $_{\pm  4.2}$}} &  \textcolor{ForestGreen}{\textbf{86.8 $_{\pm  4.4}$}} &  85.4 $_{\pm  3.3}$ & 85.9 $_{\pm  4.4}$ & 85.8 $_{\pm  3.1}$ & 85.7 $_{\pm  5.2 }$ \\
        
        spambase & 90.4 $_{\pm  2.1}$ & 91.4 $_{\pm  1.9}$ & 87.9 $_{\pm  2.4}$ & 91.9 $_{\pm  1.9}$ & \textcolor{Mulberry}{\textbf{93.3 $_{\pm  2.0}$}} &  92.5 $_{\pm  2.6}$ & \textcolor{ForestGreen}{\textbf{94.9 $_{\pm  1.2}$}} &  89.8 $_{\pm  2.8}$ & \textcolor{Bittersweet}{\textbf{94.6 $_{\pm  1.5}$}} &  92.5 $_{\pm  2.0}$ & 92.9 $_{\pm  1.3 }$ \\
        
        credit-g & 63.3 $_{\pm  6.3}$ & 65.2 $_{\pm  5.7}$ & 58.5 $_{\pm  3.7}$ & 64.2 $_{\pm  8.2}$ & 63.4 $_{\pm  5.6}$ & 63.6 $_{\pm  7.5}$ & \textcolor{Mulberry}{\textbf{66.8 $_{\pm  4.9}$}} &  62.1 $_{\pm  5.4}$ & \textcolor{Bittersweet}{\textbf{67.7 $_{\pm  4.7}$}} &  \textcolor{ForestGreen}{\textbf{68.8 $_{\pm  5.4}$}}  & 65.7 $_{\pm  5.0 }$ \\
        
        friedman & \textcolor{Bittersweet}{\textbf{84.9 $_{\pm  6.7}$}} &  69.8 $_{\pm  4.7}$ & 80.4 $_{\pm  6.6}$ & 61.2 $_{\pm  7.5}$ & 77.8 $_{\pm  5.3}$ & 68.5 $_{\pm  5.8}$ & \textcolor{ForestGreen}{\textbf{86.4 $_{\pm  8.7}$}} &  \textcolor{Mulberry}{\textbf{82.0 $_{\pm  4.8}$}} &  61.5 $_{\pm  8.3}$ & 59.8 $_{\pm  8.4}$ & 80.0 $_{\pm  7.4 }$ \\
        
        usps & 93.4 $_{\pm  2.9}$ & \textcolor{ForestGreen}{\textbf{98.5 $_{\pm  1.2}$}} &  92.3 $_{\pm  1.6}$ & 96.6 $_{\pm  1.4}$ & 97.2 $_{\pm  1.4}$ & 97.5 $_{\pm  1.2}$ & \textcolor{Bittersweet}{\textbf{98.0 $_{\pm  1.5}$}} &  93.5 $_{\pm  2.2}$ & \textcolor{Mulberry}{\textbf{97.8 $_{\pm  1.7}$}} &  \textcolor{Bittersweet}{\textbf{98.0 $_{\pm  1.1}$}} &  97.5 $_{\pm  1.6 }$ \\
        
        bioresponse & 71.8 $_{\pm  1.5}$ & 71.3 $_{\pm  3.2}$ & 71.0 $_{\pm  2.1}$ & 72.3 $_{\pm  2.2}$ & \textcolor{ForestGreen}{\textbf{78.9 $_{\pm  2.1}$}} &  \textcolor{Bittersweet}{\textbf{78.1 $_{\pm  2.0}$}} &  \textcolor{Mulberry}{\textbf{77.7 $_{\pm  3.3}$}} &  72.6 $_{\pm  2.9}$ & \textcolor{Bittersweet}{\textbf{78.1 $_{\pm  2.6}$}} &  74.6 $_{\pm  3.0}$ & 72.9 $_{\pm  2.6 }$ \\
        
        speeddating & 63.8 $_{\pm  3.3}$ & 64.0 $_{\pm  1.1}$ & 60.1 $_{\pm  3.0}$ & 58.9 $_{\pm  9.3}$ & 58.0 $_{\pm  1.5}$ & 57.4 $_{\pm  1.7}$ & \textcolor{Bittersweet}{\textbf{69.1 $_{\pm  1.9}$}} &  65.5 $_{\pm  2.8}$ & \textcolor{Mulberry}{\textbf{65.8 $_{\pm  2.1}$}} &  \textcolor{ForestGreen}{\textbf{71.2 $_{\pm  2.7}$}} &  65.4 $_{\pm  1.9 }$ \\
    
        \midrule

        Avg. rank & 5.3 & 5.5 & 3.4 & 4.3 & 6.1 & 5.3 & \textcolor{Bittersweet}{\textbf{8.5}} & 4.5 & \textcolor{Mulberry}{\textbf{7.8}} & 6.5 & \textcolor{ForestGreen}{\textbf{8.8}} \\
        \bottomrule
    \end{tabular}
    }
    \label{tab__performance}
\end{table*}

\subsection{Experimental setup}

\paragraph{Datasets} Our choice of datasets is guided by a recent study about tabular data~\cite{mcelfresh_2023_tabzilla}. We choose 22 binary classification datasets with varying numbers of features and samples from OpenML. We first split the data according to 10 train/test folds provided by OpenML, which enables the comparison of our results on the test data with other works. Furthermore, we split each training fold into training and validation sets, using the same splits as~\cite{mcelfresh_2023_tabzilla}. Refer to Appendix~\ref{sec:appendix-experiments:datasets} for further details on the datasets (Table~\ref{tab__datasets}) and preprocessing.

\paragraph{Baselines} We compare \mbox{RO-FIGS} to several tree-based methods, including: decision tree~\cite{cart_1984}, random forest~\cite{breiman_2001}, extremely randomised or extra trees (ETC)~\cite{geurts_2006_extra_trees}, FIGS~\cite{tan_2022_figs}, oblique decision tree (ODT)~\cite{stepisnik_2021_oblique_pcts} and ensemble of oblique decision trees (Ens-ODT)~\cite{stepisnik_2021_oblique_pcts}, optimal tree (OT)~\cite{optimal_tree}, model tree (MT)~\cite{linear_tree}, as well as several gradient-boosted ensembles (LightGBM~\cite{ke_2017_lightgbm}, XGBoost~\cite{xgboost} and CatBoost~\cite{catboost}). We present only the results of CatBoost as the top-performing GBDT baseline. Further results on the performance of the other two are given in Appendix~\ref{sec:appendix-results:baselines} (Table~\ref{tab__gbdts_performance}). We also include an MLP with two hidden layers as a neural network baseline. Note that we also compared \mbox{RO-FIGS} to TabPFN~\cite{hollmann_2023_tabpfn}, a recently proposed transformer-based approach tailored for tabular data tasks. However, since TabPFN has been pre-trained on some of the datasets we use for evaluation, we include this analysis in Appendix~\ref{sec:appendix-results:baselines} (Table~\ref{tab__tabpfn}), as a direct comparison would be biased due to data leakage.

\paragraph{Hyperparameter optimisation} 
For tuning RO-FIGS, we perform a grid search over the parameters that control the impurity decrease and the size of the beam. More details can be found in Appendix~\ref{sec:appendix-experiments:hyperparameter-tuning}. We use hyperopt~\cite{hyperopt} to tune the hyperparameters of the baselines. The configuration, which leads to the best performance on validation data, is then used to train final models on the joint training data (comprising training and validation sets), and their performance on the hold-out test set is reported. The exception is the MLP baseline, which requires validation data for early stopping. We use 30 iterations of tuning for all methods except model trees, which we limit to five iterations due to a high computational overhead. \looseness -1

\paragraph{Metrics} To evaluate performance, we measure the balanced accuracy on the test split. We perform a 10-fold cross-validation and report the mean \mbox{$\pm$} standard deviation, averaged across 10 folds. We also report the mean rank of each method, ranked according to their performance, with higher ranks denoting better performance. To demonstrate the compactness of \mbox{RO-FIGS} models, we follow Tan et al.~\cite{tan_2022_figs} and compute the number of splits and trees, which allows us to make further comparisons among tree-based models. These two reflect the model complexity and thus implicitly measure interpretability, as we assume more compact models to be more interpretable. Note that we do not explicitly analyse the depth of the trees, as we found it to be unreliable for measuring model complexity given different tree-learning methods. As a measure of efficiency, we also assess the balance between performance and compactness of tree-based models. Finally, we analyse oblique splits in \mbox{RO-FIGS} models to demonstrate their expressiveness. By identifying combinations of features that often appear together in the splits, we show which feature interactions positively contribute to the model's performance. We analyse this data along with SHAP plots, which provide post-hoc model explainability.

\subsection{How accurate are \mbox{RO-FIGS} models?}
\label{sec:accuracy}

Table~\ref{tab__performance} shows that \mbox{RO-FIGS}, in general, is able to outperform all baselines. It consistently dominates on smaller datasets, performing comparably to other tree-based ensembles on datasets with a larger number of samples or features. The neural network baseline, MLP, is competitive, achieving top performance on six datasets, but its rank is diminished due to poor performance on a few datasets (e.g., \textit{pc1} and \textit{friedman}). 

\begin{figure}[tb]
    \centering
    \includegraphics[width=0.5\textwidth]{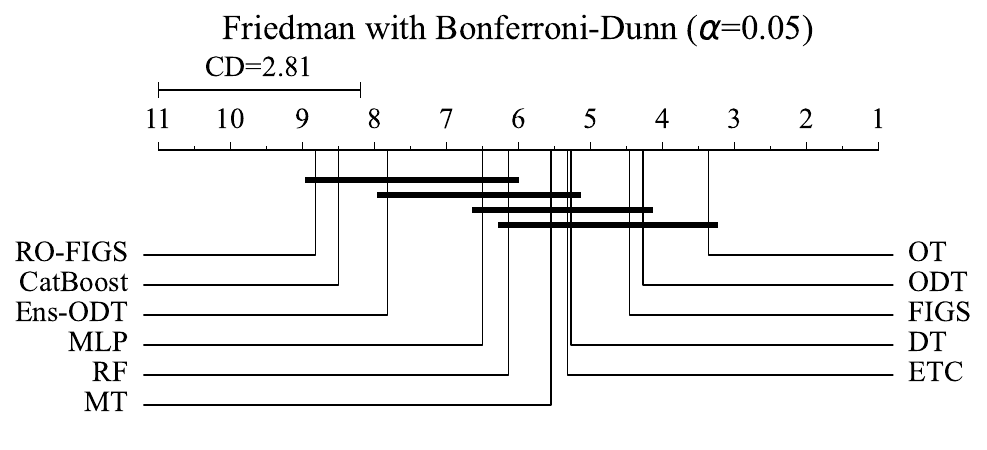}
    \caption{Comparison of the average ranks of model performance, comparing RO-FIGS against the baselines, with respect to the Friedman statistical test followed by the Bonferroni-Dunn post-hoc test (p $<$ 0.05). RO-FIGS ranks best (higher values are better), and statistically significantly better (outside the $\mathit{CD}$ interval) than FIGS and five other baseline methods: MT, ETC, DT, ODT and OT.} 
    \label{fig__cd_diagram} 
\end{figure}

To further assess the performance differences of the evaluated methods, we follow a standard statistical procedure recommended by Demšar~\cite{demsar_2006}. We employ the corrected~\cite{ImanDevenport} Friedman statistical test~\cite{friedman_1940} to determine the performance difference between \mbox{RO-FIGS} and the baselines. This is followed by the Bonferroni-Dunn~\cite{dunn_1961} post-hoc test to identify which differences are statistically significant. Fig.~\ref{fig__cd_diagram} depicts the critical difference diagram (with a significance level threshold set at $95\%, p = 0.05$). \mbox{RO-FIGS} ranks best amongst all methods (higher values are better), and statistically significantly better than FIGS as well as other baselines outside the marked interval (for critical distance $\mathit{CD}=2.81$), which includes MT, ETC, DT, ODT, and OT. Note that, while \mbox{RO-FIGS} ranks better than CatBoost, Ens-ODT, MLP, and RF, this difference is not statistically significant.

\textbf{Oblique splits in \mbox{RO-FIGS} models are beneficial for performance.} We attribute this to oblique splits not being restricted to splitting along a single feature dimension. This flexibility allows them to capture more intricate, non-linear relationships between features, leading to better-performing models. We observe that, with the exception of two datasets, \mbox{RO-FIGS} always outperforms FIGS, suggesting that while both methods benefit from the additive structure, \mbox{RO-FIGS's} superior performance stems from using oblique splits. This comparison and further study on how varying the size of the beam (\textsc{beam\_size} parameter) influences the performance of \mbox{RO-FIGS} models is discussed in Appendix~\ref{sec:appendix-results:ablations}.

\subsection{How efficient are \mbox{RO-FIGS} models?}
\label{sec:efficiency}

\begin{figure*}[th]
    \centering
    \begin{minipage}[b]{0.32\textwidth}
        \centering
        \includegraphics[width=\textwidth]{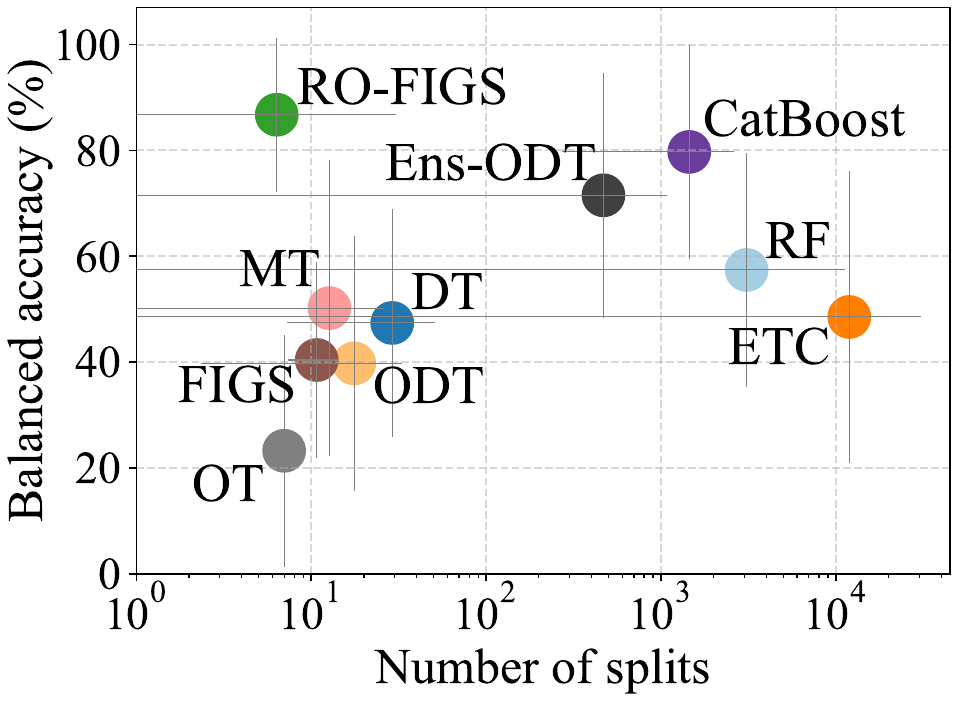}
    \end{minipage}
    \hfill
    \begin{minipage}[b]{0.32\textwidth}
        \centering
        \includegraphics[width=\textwidth]{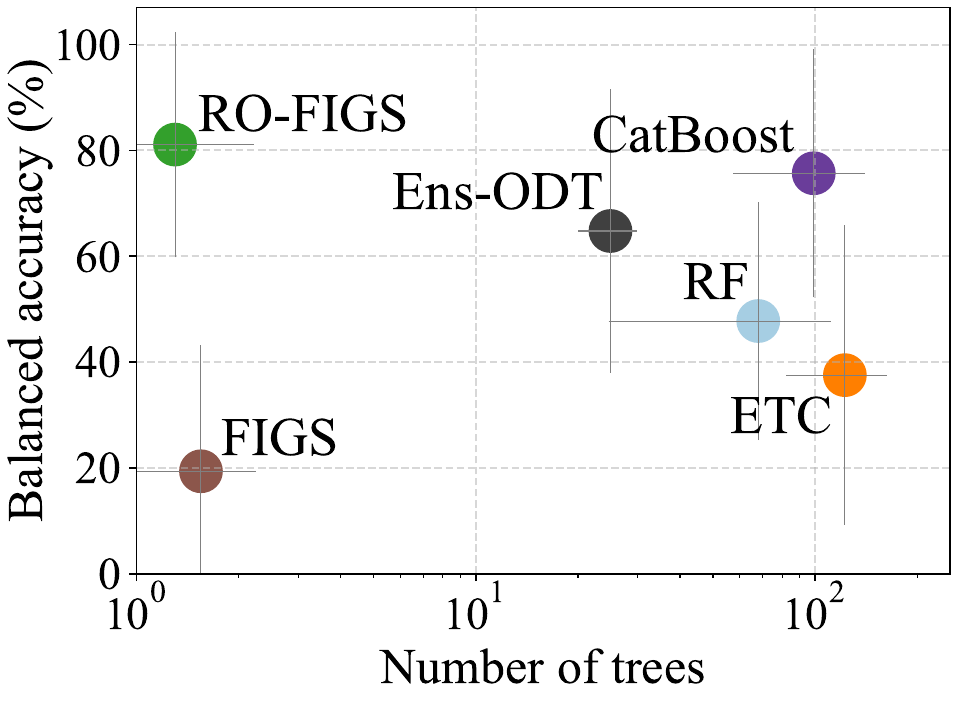}
    \end{minipage}
    \hfill
    \begin{minipage}[b]{0.32\textwidth}
        \centering
        \includegraphics[width=\textwidth]{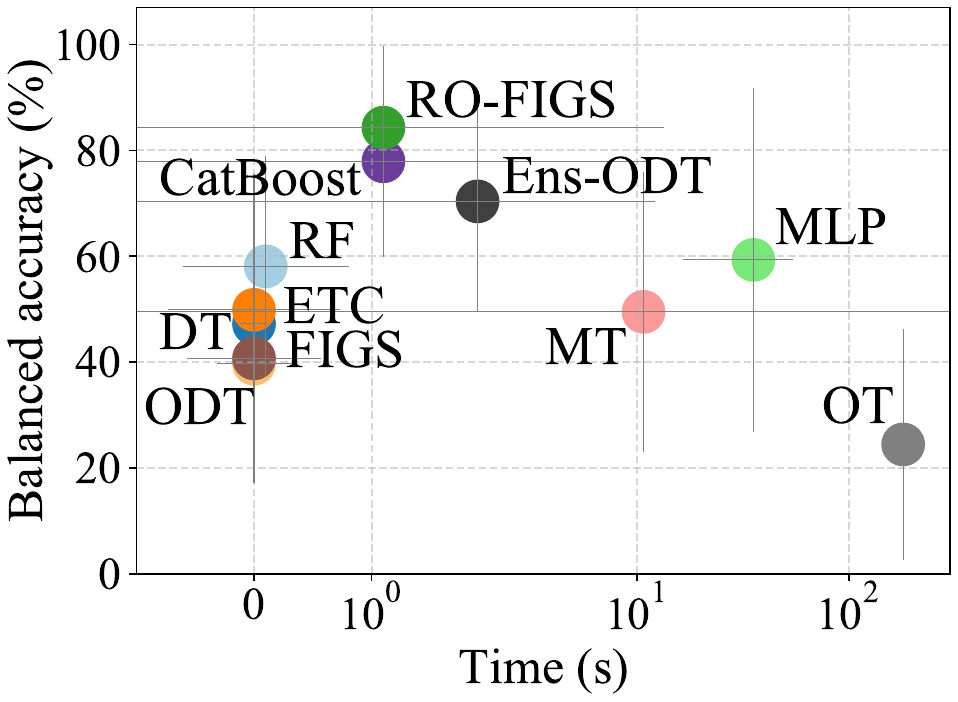}
    \end{minipage}
    \caption{Left and middle: Comparison between the performance and the number of splits and trees in tree-based models. RO-FIGS models achieve the best balance, as they are well-performing and small in size. Right: Comparison between the performance and training time. RO-FIGS offers a good balance between performance and computational cost. Performance is calculated as mean normalised accuracy for each method, over all datasets, with error bars spanning the 20th/80th percentile over all datasets. Note the logarithmic scale on the x-axes and the omission of single-tree methods in the middle figure. 
}
    \label{fig__split_trees_plot} 
\end{figure*}

We assess efficiency as the balance between performance and compactness of the models. Since we cannot directly compare the compactness of tree- and network-based models, we exclude the MLP baseline from this analysis. We acknowledge, however, that neural networks have millions of parameters and thus offer very limited interpretability for their decisions. 

We first use two quantitative measures to assess the size of \mbox{RO-FIGS} models: the number of trees and the total number of splits across all trees. We then analyse the balance between each measure and the mean accuracy. This analysis provides an overview of potential trade-offs between compactness and performance, which is crucial in practical applications.

\textbf{\mbox{RO-FIGS} produces performant small-sized models.} Fig.~\ref{fig__split_trees_plot} (left and middle) shows a comparison between the size of tree-based models and their performance. The ideal, most efficient model would be positioned in the top-left corner, achieving the best accuracy whilst being compact. It is evident that \mbox{RO-FIGS} offers the best balance: it builds small and therefore more interpretable models, whilst outperforming the baselines. While \mbox{RO-FIGS} and CatBoost (which ranks second to our method) achieve similar average performance, the former builds much smaller models and is, therefore, more suitable for applications where simpler, more interpretable models are desired~\cite{rudin_2019}. Further details are provided in Appendix~\ref{sec:appendix-results:efficiency}.\looseness-1 

\mbox{RO-FIGS} is able to take advantage of the additive structure of the data, if that exists, and build multiple trees in summation. Our experiments show that it generates compact, yet high-performing, models. They consist of less than five trees on average, which is comparable to that of FIGS and much lower than ensembles (see Table~\ref{tab__trees} in Appendix~\ref{sec:appendix-results:efficiency}). Moreover, we observe that both FIGS and \mbox{RO-FIGS} build either exactly one tree or multiple trees on most datasets. This confirms that both methods uncover the additive nature of the data if present. Overall, \mbox{RO-FIGS} builds models with a small number of trees.

With regards to the number of splits, \mbox{RO-FIGS} builds models with fewer than 15 splits on 15 datasets (see Table~\ref{tab__splits} in Appendix~\ref{sec:appendix-results:efficiency}). Furthermore, on 10 of these datasets, they have fewer than five splits. As with trees, the number of splits is comparable to FIGS, with few exceptions. Note that, in our analysis, we limited the maximum number of splits in FIGS to 20, as in~\cite{tan_2022_figs}, so it is likely that other methods will create models with more splits, especially on harder tasks. \mbox{RO-FIGS} models have even fewer splits than single decision trees on more than half of the datasets while performing much better. Tree ensembles, as expected, exhibit orders of magnitude more splits than a single tree- and FIGS-based models. Overall, \mbox{RO-FIGS} builds models with a low number of splits. 

We have shown that \mbox{RO-FIGS} leads to good and stable performance with compact models. However, there is a trade-off between predictive performance and computational cost for training the models (see Fig.~\ref{fig__split_trees_plot} (right)). Compared to other ensemble baselines, \mbox{RO-FIGS} may require more training time due to its on-the-fly learning and evaluation of oblique splits. Therefore, in scenarios with high-dimensional datasets, \mbox{RO-FIGS} can scale poorly with the number of potential splits and the size of the beam \textsc{beam\_size} (i.e., size of the feature sets). More specifically, in the worst case, the complexity of \mbox{RO-FIGS} is $\mathcal{O}(irm^2n^2d)$, where $m, n, d, i$, and $r$ denote the number of splits, samples, features, gradient descent iterations (100 by default), and repetitions (five by default), respectively. \mbox{RO-FIGS} depends on the optimisation procedure for learning the splits, typically using a subset of $d$ (i.e., \textsc{beam\_size}) features at every split. This process could be improved by computing potential splits within each iteration in parallel and using more effective heuristics or domain-specific knowledge for feature selection. Nevertheless, \mbox{RO-FIGS} still offers a favourable balance between practical time complexity and performance.

\subsection{How expressive are \mbox{RO-FIGS} models?}
\label{sec:expressiveness}

To demonstrate the expressiveness of \mbox{RO-FIGS} models, we now analyse their oblique splits. First, we inspect the average number of features per split, which is constrained by the size of the beam (\textsc{beam\_size}). Due to the L$\frac{1}{2}$ norm in \eqref{eq:split_optimisation}, the number of features with non-zero weights is usually much lower than this upper limit, keeping the size of individual splits relatively small (see Table~\ref{tab__grid_rofigs} in Appendix~\ref{sec:appendix-experiments:hyperparameter-tuning} for more details). A low number of features decreases the complexity of individual splits, making them more comprehensible and interpretable. This behaviour also suggests that \mbox{RO-FIGS} can, like other tree-based methods, recognise more informative features.   

\begin{figure}[b!]
    \centering
    \includegraphics[width=0.39\textwidth]{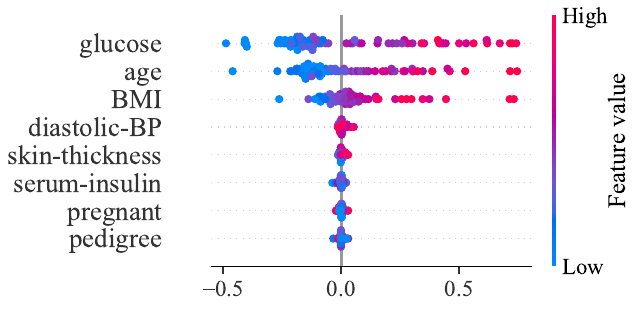}
    \caption{SHAP summary plot of \mbox{RO-FIGS} on the \textit{diabetes} dataset (fold 0). Similar to the baselines, \textit{glucose}, \textit{age}, and \textit{BMI} features are contributing to the model the most. \mbox{RO-FIGS} with the accuracy of 73.6\% (see Table~\ref{tab__performance}) outperforms all baselines on this dataset with only one split, demonstrating that a linear combination of features \textit{glucose}, \textit{age}, and \textit{BMI} is optimal.}
    \label{fig__rofigs_diabetes_only}
\end{figure}

Next, we showcase the expressiveness of \mbox{RO-FIGS} models by analysing their splits on the \textit{diabetes} dataset with eight features. Across all folds, \mbox{RO-FIGS} models consist of one tree with one split. Moreover, all splits consist of a linear combination of three features: \textit{glucose}, \textit{age}, and \textit{BMI}. This is consistent with the SHAP summary plot of the \mbox{RO-FIGS} model (on the first fold), shown in Fig.~\ref{fig__rofigs_diabetes_only}, which shows that only these features contribute to the model's predictions. For comparison, SHAP plots of the baselines also identify these three features as very important. However, unlike other methods, RO-FIGS has the unique advantage of inherently capturing feature interactions within splits (as they are defined as linear combinations of features, $\Phi$), thereby providing insights into specific interactions between features. More details can be found in Appendix~\ref{sec:appendix-results:expressiveness}.

Overall, \mbox{RO-FIGS} consistently builds splits with multiple features and helps uncover important relationships between features. These results demonstrate that splits in \mbox{RO-FIGS} are more expressive and informative than univariate splits while considering a reasonable number of features at once. 
Furthermore, \mbox{RO-FIGS} models can provide additional information about the feature interactions that cannot be learnt from analysing splits of tree-based models with univariate splits or post-hoc explanation tools such as SHAP alone. 

\subsection{Limitations}

\mbox{RO-FIGS} demonstrates good predictive performance while producing compact and interpretable models. Nevertheless, we acknowledged a few limitations related to its scalability to larger datasets and versatility to mixed-type tabular datasets. Future work includes exploring more efficient approaches for feature construction and selection, which will serve as the basis for learning oblique trees from numerical and nominal features, but also for reducing the computational burden for learning performant RO-FIGS.

\section{Conclusion}
\label{sec:conclusion}

In this work, we introduced \mbox{RO-FIGS}, a tree-based method for tabular data. 
\mbox{RO-FIGS} builds compact models with multiple trees by using residuals of previous trees. Its splits are oblique and expressed as linear combinations of multiple features, learnt from random subsets of features. We show that \mbox{RO-FIGS} achieves the highest average performance, surpassing both traditional tree-based methods and 
MLP, whilst building much smaller models. Due to the nature of oblique splits in \mbox{RO-FIGS}, we provide insights into the interactions between features that contribute to the models' performance. We show that this information enriches insights from SHAP summary plots, enhancing the interpretability of \mbox{RO-FIGS}. Although we focus on classification tasks in this work, the method is also applicable to regression tasks. The proposed method is especially well-suited for real-world applications, where the balance between accuracy and interpretability is crucial. \looseness -1

\clearpage
\bibliographystyle{IEEEtran}
\bibliography{IEEEabrv,main}

\clearpage
\pagebreak

\begin{appendices}

\onecolumn
\begin{center}
{\bf
\sffamily\normalfont
\huge{Supplementary material for \\ 
\vspace*{2mm}

\Huge RO-FIGS: Efficient and Expressive Tree-Based Ensembles for Tabular Data}}
\end{center}

\vspace*{4mm}

\begin{center}
{\sublargesize Ur\v{s}ka Matja\v{s}ec$^1$\quad\quad\quad\quad\quad Nikola Simidjievski$^{2,1}$\quad\quad\quad\quad\quad Mateja Jamnik$^1$}\\

{$^1$\textit{Department of Computer Science and Technology}, $^2$PBCI, \textit{Department of Oncology}\\
\textit{University of Cambridge, Cambridge, UK}\\
\{um234,ns779,mj201\}@cam.ac.uk}
\end{center}

\vspace*{4mm}

\section{Extended experimental setup}
\label{sec:appendix-experiments}

Here, we give more details about the datasets used for evaluation (see Table~\ref{tab__datasets}), discuss different encodings of nominal features in \mbox{RO-FIGS} (see Table~\ref{tab__encodings}), and provide details about the search space for hyperparameter tuning of the tree-based baselines (see Table~\ref{tab__hyperopt_baselines}) and grid search space along with the best per-dataset configuration for \mbox{RO-FIGS} (see Table~\ref{tab__grid_rofigs}).

\subsection{Datasets}
\label{sec:appendix-experiments:datasets}

All datasets are publicly available on OpenML and summarised in Table~\ref{tab__datasets}. We first split the data according to 10 train/test folds provided by OpenML and then split each training fold into training and validation sets, using the same splits as McElfresh et al.~\cite{mcelfresh_2023_tabzilla}. Each dataset is therefore split into training (80\%), validation (10\%), and test (10\%) sets. 

Most baselines expect numerical input, therefore we one-hot encode nominal features. The only exceptions are \mbox{CatBoost} and \mbox{LightGBM}, to which we pass the indices of nominal features. For \mbox{RO-FIGS}, we evaluate three encoding approaches of transforming nominal features into numerical: $\textsc{E-ohe}$, $\textsc{E-count}$, and $\textsc{E-prop}$, as computing linear combinations of nominal features is not trivial. Table~\ref{tab__encodings} demonstrates that one-hot encoding outperforms the other two, which transform nominal features into ordered ones without increasing the dimensionality of the data. Therefore, we use one-hot encoding to transform nominal features into numerical ones.

The data is min-max scaled to [0, 1]. Specifically, during hyperparameter tuning (with models trained on the training data and evaluated on the validation data), the scaling transformation is learnt from the training data and applied to the validation data. When training the final model on the combined training and validation data (and evaluated on test data), the transformation learnt from the joint dataset is applied to the test data.

\subsection{Hyperparameter tuning}
\label{sec:appendix-experiments:hyperparameter-tuning}

\paragraph{Baselines}

We employ \texttt{hyperopt}~\cite{hyperopt} with the tree-structured Parzen estimator to fine-tune hyperparameters for the tree-based baselines. Each baseline undergoes 30 tuning iterations, with the exception of the model tree, which is limited to five iterations due to its high computational cost. The details about the search space of hyperparameters are shown in Table~\ref{tab__hyperopt_baselines}. TabPFN requires no tuning and we use its default hyperparameter values. For optimal trees, we set the depth to three and trained each model for up to three minutes.

Because we compare \mbox{RO-FIGS} to the baselines w.r.t.\ the number of trees and splits, we additionally introduce the following penalties to promote the compactness in the tree-based baselines:
\begin{itemize}
    \item \textbf{DT}: 0.001 $\times$ max\_depth
    \item \textbf{MT} using \texttt{RidgeClassifier} as the base estimator: 0.001 $\times$ max\_depth
    \item \textbf{ODT}: 0.001 $\times$ max\_depth
    \item \textbf{RF, ETC, CatBoost, LightGBM, XGBoost}: 0.001 $\times$ max\_depth + 0.0001 $\times$ n\_estimators
    \item \textbf{FIGS}: 0.001 $\times$ max\_rules + 0.001 $\times$ max\_trees
    \item \textbf{Ens-ODT}: 0.001 $\times$ max\_depth + 0.0001 $\times$ num\_trees
\end{itemize}

\paragraph{RO-FIGS}

To better understand the behaviour of \mbox{RO-FIGS} models across a spectrum of hyperparameter values, we perform a grid search over two hyperparameters: \textsc{beam\_size}, which limits the number of features considered per split, and \textsc{min\_imp\_dec}, defining the stopping condition for training. We use a constant search space for the minimum impurity decrease parameter and tailor the other parameter to the number of features in the dataset (i.e., d):
\begin{itemize}
    \item \textsc{min\_imp\_dec}: 0.05, 0.1, 1, 5, 10, and
    \item \textsc{beam\_size}: 1, 2, $\sqrt{d}$, $\frac{d}{4}$, $\frac{d}{2}$, d; where d is the total number of features in a dataset.
\end{itemize}

The middle two columns in Table~\ref{tab__grid_rofigs} show the optimal configuration, which leads to the best performance on the test data. It should be noted that the \textsc{beam\_size} value is the maximum number of features that \mbox{RO-FIGS} considers at each iteration. In practice, only a small proportion of features have non-zero weights, leading to a much lower average number of features per split. This is reported in the last column of Table~\ref{tab__grid_rofigs}.

\begin{table*}[ht]
\caption{Properties of the benchmark datasets for binary classification tasks. \textit{Nominal} indicates the number of nominal categorical features, while \textit{Final} shows the total number of features after nominal features have been one-hot encoded (datasets are ordered w.r.t. the latter).}
\centering
\begin{tabular}{llrrrr}
    \toprule
    Dataset & Original name (OpenML id) & Numerical & Nominal & Final & Samples \\
    \midrule
        blood & blood-transfusion-service-center (1464) & 4 & 0 & 4 &  748 \\ 
        diabetes & diabetes (37) & 8 & 0 & 8 &  768 \\ 
        breast-w & breast-w (15) & 9 & 0 & 9 &  699 \\ 
        ilpd & ilpd (1480) & 9 & 1 & 11 &  583 \\ 
        monks2 & monks-problems-2 (334) & 0 & 6 & 17 &  601 \\ 
        climate & climate-model-simulation-crashes (40994) & 18 & 0 & 18 &  540 \\ 
        kc2 & kc2 (1063) & 21 & 0 & 21 &  522 \\ 
        pc1 & pc1 (1068) & 21 & 0 & 21 &  1,109 \\ 
        kc1 & kc1 (1067) & 21 & 0 & 21 &  2,109 \\ 
        heart & heart-c (982) & 6 & 7 & 26 &  303 \\ 
        tictactoe & tic-tac-toe (50) & 0 & 9 & 27 &  958 \\ 
        wdbc & wdbc (1510) & 30 & 0 & 30 &  569 \\ 
        churn & churn (40701) & 16 & 4 & 33 &  5,000 \\ 
        pc3 & pc3 (1050) & 37 & 0 & 37 &  1,563 \\ 
        biodeg & qsar-biodeg (1494) & 41 & 0 & 41 &  1,055 \\ 
        credit & credit-approval (29) & 6 & 9 & 51 &  690 \\ 
        spambase & spambase (44) & 57 & 0 & 57 &  4,601 \\ 
        credit-g & credit-g (31) & 7 & 13 & 61 &  1,000 \\ 
        friedman & fri\_c4\_500\_100 (610) & 100 & 0 & 100 &  500 \\ 
        usps & USPS (41964) & 256 & 0 & 256 &  1,424 \\ 
        bioresponse & Bioresponse (45019) & 419 & 0 & 419 &  3,434 \\ 
        speeddating & SpeedDating (40536) & 59 & 61 & 503 &  8,378 \\ 
    \bottomrule
\end{tabular}
\label{tab__datasets}
\end{table*}
\begin{table*}[th!]
\caption{Classification performance of RO-FIGS on datasets that include nominal features, using three different encodings for nominal features: $\textsc{E-ohe}$, $\textsc{E-count}$ and $\textsc{E-prop}$. We report mean $\pm$ std of the test balanced accuracy across 10 folds. One-hot encoding ($\textsc{E-ohe}$) leads to the best performance on average.}
\centering
\begin{tabular}{lrrr}
    \toprule
    Dataset & $\textsc{E-ohe}$ & $\textsc{E-count}$ & $\textsc{E-prop}$  \\
    \midrule
    ilpd & 61.9 $\pm$ 10.4 & 62.9 $\pm$ 7.7 & 61.6 $\pm$ 4.2 \\
    monks-2 & 99.5 $\pm$ 0.8 & 76.5 $\pm$ 6.6 & 76.2 $\pm$ 4.5 \\
    heart & 84.8 $\pm$ 6.4 & 83.5 $\pm$ 6.7 & 81.8 $\pm$ 5.4 \\
    tictactoe & 94.4 $\pm$ 3.1 & 78.1 $\pm$ 4.7 & 87.5 $\pm$ 3.9 \\
    churn & 86.3 $\pm$ 2.0 & 82.3 $\pm$ 3.6 & 86.7 $\pm$ 2.1 \\
    credit & 85.7 $\pm$ 5.2 & 86.7 $\pm$ 3.8 & 86.2 $\pm$ 3.1 \\
    credit-g & 65.7 $\pm$ 5.0 & 65.6 $\pm$ 5.2 & 66.4 $\pm$ 2.3 \\
    \midrule
    Avg. rank & 1.7 & 2.1 & 2.1 \\
    \bottomrule
\end{tabular}
\label{tab__encodings}
\end{table*}
\begin{table*}[h]
    \caption{Search space for hyperparameters of tree-based baselines. We use \texttt{hp.quniform} to define the space unless marked otherwise. Symbols $*$, $\dagger$, and $\star$ denote the use of \texttt{hp.uniform}, \texttt{hp.loguniform}, and \texttt{hp.choice}, respectively. $d$ denotes the total number of features per dataset.}
    \centering
    \begin{tabular}{llll}
        \toprule
        Method & Hyperparameter & Type & Range \\
        \midrule
        DT          & max\_depth                & Discrete      & [1, \dots, 30] \\
                    & min\_samples\_leaf        & Discrete      & [1, \dots, 40] \\
                    & min\_samples\_split       & Discrete      & [2, \dots, 40] \\ \midrule
        MT          & alpha $*$                 & Continuous    & [0.5, \dots, 1] \\
                    & max\_depth                & Discrete      & [1, \dots, 20] \\
                    & min\_samples\_leaf        & Discrete      & [3, \dots, 40] \\
                    & min\_samples\_split       & Discrete      & [6, \dots, 40] \\ \midrule
        ODT         & max\_depth                & Discrete      & [1, \dots, 30] \\
                    & max\_features $\star$     & Choice        & $[1, 2, \sqrt{d}, \frac{d}{4}, \frac{d}{2}, d]$ \\
                    & min\_examples\_to\_split  & Discrete      & [2, \dots, 40] \\ \midrule
        RF          & n\_estimators             & Discrete      & [5, \dots, 250] \\
                    & max\_depth                & Discrete      & [1, \dots, 30] \\
                    & min\_samples\_leaf        & Discrete      & [1, \dots, 40] \\
                    & min\_samples\_split       & Discrete      & [2, \dots, 40] \\ \midrule
        ETC         & n\_estimators             & Discrete      & [5, \dots, 250] \\
                    & max\_depth                & Discrete      & [1, \dots, 30] \\
                    & min\_samples\_leaf        & Discrete      & [1, \dots, 40] \\
                    & min\_samples\_split       & Discrete      & [2, \dots, 40] \\ \midrule
        CatBoost    & n\_estimators             & Discrete      & [5, \dots, 250] \\
                    & max\_depth                & Discrete      & [1, \dots, 30] \\
                    & learning\_rate $\dagger$  & Continuous    & [0.01, \dots, 0.25] \\
                    & num\_leaves               & Discrete      & [2, \dots, 40] \\
                    & min\_child\_samples       & Discrete      & [1, \dots, 40] \\ \midrule
        LightGBM    & n\_estimators             & Discrete      & [5, \dots, 250] \\
                    & max\_depth                & Discrete      & [1, \dots, 30] \\
                    & learning\_rate $\dagger$  & Continuous    & [0.01, \dots, 0.25] \\
                    & num\_leaves               & Discrete      & [2, \dots, 40] \\
                    & min\_child\_samples       & Discrete      & [1, \dots, 40] \\ \midrule
        XGBoost     & n\_estimators             & Discrete      & [5, \dots, 250] \\
                    & max\_depth                & Discrete      & [1, \dots, 30] \\
                    & learning\_rate $\dagger$  & Continuous    & [0.01, \dots, 0.25] \\
                    & min\_split\_loss          & Discrete      & [0, \dots, 40] \\
                    & min\_child\_weight        & Discrete      & [1, \dots, 40] \\ \midrule
        FIGS        & max\_rules                & Discrete      & [1, \dots, 20] \\
                    & max\_trees                & Discrete      & [1, \dots, 20] \\ \midrule
        
        EnsODT         & max\_depth                & Discrete      & [1, \dots, 30] \\
                    & max\_features $\star$     & Choice        & $[1, 2, \sqrt{d}, \frac{d}{4}, \frac{d}{2}, d]$ \\
                    & min\_examples\_to\_split  & Discrete      & [2, \dots, 40] \\ 
                    & num\_trees                & Discrete      & [5, \dots, 50] \\ \midrule
        
        MLP         & num\_epoch                & Discrete      & [1,000] \\
                    & patience                  & Discrete      & [50, \dots, 300] \\
                    & learning\_rate $\dagger$  & Continuous    & [0.0001, \dots, 0.01] \\
                    & dropout\_rate $*$         & Continuous    & [0, \dots, 0.25] \\
                    & weight\_decay $*$         & Continuous    & [0, \dots, 0.01] \\
                    & L1 $*$                    & Continuous    & [0, \dots, 0.01] \\
                    
        \bottomrule
    \end{tabular}
    \label{tab__hyperopt_baselines}
\end{table*}

\begin{table*}[h]
    \caption{Optimal configuration, leading to the best performance on test data, is reported in the middle two columns: \textsc{min\_imp\_dec} is a minimum impurity decrease value used as a stopping condition, and \textsc{beam\_size} represents the number of features that are considered per split. Although the optimal \textsc{beam\_size} value is often large, the actual average number of features per split, reported in the last column, is much lower.}
    \centering
    \begin{tabular}{lccc}
        \toprule
                & \multicolumn{2}{c}{Optimal} & \multicolumn{1}{c}{Actual} \\
        Dataset & \textsc{min\_imp\_dec} & \textsc{beam\_size} & \multicolumn{1}{c}{\parbox{2.3cm}{\#features per split}} \\

        \midrule
            blood           & 5 & 2 &  1.6 $\pm$ 0.2 \\
            diabetes        & 10 & 8 &  3.0 $\pm$ 0.0 \\
            breast-w        & 1 & 9 &  4.1 $\pm$ 0.4 \\
            ilpd            & 0.1 & 5 &  2.0 $\pm$ 0.0 \\
            monks2          & 0.1 & 17 &  7.2 $\pm$ 0.5 \\
            climate         & 0.1 & 18 &  7.7 $\pm$ 0.6 \\
            kc2             & 10 & 21 &  10.7 $\pm$ 0.8 \\
            pc1             & 0.1 & 4 &  1.7 $\pm$ 0.1 \\
            kc1             & 0.05 & 4 &  2.0 $\pm$ 0.1 \\
            heart           & 10 & 26 &  9.7 $\pm$ 0.7 \\
            tictactoe       & 0.05 & 13 &  4.5 $\pm$ 0.2 \\
            wdbc            & 1 & 30 &  10.4 $\pm$ 1.4 \\
            churn           & 5 & 33 &  9.6 $\pm$ 0.9 \\
            pc3             & 0.1 & 18 &  7.4 $\pm$ 0.2 \\
            biodeg          & 10 & 41 &  14.6 $\pm$ 1.7 \\
            credit          & 1 & 12 &  3.9 $\pm$ 0.5 \\
            spambase        & 0.1 & 57 &  31.4 $\pm$ 0.5 \\
            credit-g        & 1 & 2 &  1.2 $\pm$ 0.1 \\
            friedman        & 10 & 25 &  9.1 $\pm$ 1.3 \\
            usps            & 1 & 256 &  46.4 $\pm$ 15.2 \\
            bioresponse     & 0.05 & 419 &  45.8 $\pm$ 13.3 \\
            speeddating     & 10 & 125 &  50.1 $\pm$ 3.9 \\

        \bottomrule
    \end{tabular}

    \label{tab__grid_rofigs}
\end{table*}

\newpage
\clearpage
\pagebreak
\clearpage 

\section{Extended general results}
\label{sec:appendix-results}

In this section, we provide additional results from Section~\ref{sec:experiments}. 

\subsection{Performance of additional baselines}
\label{sec:appendix-results:baselines}

We first show additional results of baselines that were excluded in the paper. Table~\ref{tab__gbdts_performance} compares the performance of three gradient-boosted decision tree methods---CatBoost, LightGBM, and XGBoost---with RO-FIGS. We choose \mbox{CatBoost} as a representative method of the class of gradient-boosted decision trees because it performs the best out of three. 
Table~\ref{tab__tabpfn} then compares \mbox{RO-FIGS} and TabPFN, which we excluded as a baseline due to potential bias and misleading results; TabPFN was pre-trained on some of the datasets we used for evaluation.

\begin{table*}[htb]
    \caption{Classification performance of LightGBM, XGBoost, CatBoost, and RO-FIGS on 22 tabular datasets. We report mean $\pm$ std of the test balanced accuracy across 10 folds. The highest accuracy for each dataset across all methods is bolded, and the highest accuracy for each dataset across baselines is underlined. CatBoost is performing best out of the three gradient-boosted decision tree methods with 12 wins.}
    \centering

    \begin{tabular}{lrrr|r}
        \toprule
        Dataset & LightGBM & XGBoost & CatBoost & RO-FIGS \\
        \midrule

        blood           & \underline{60.6 $\pm$ 5.5} & 59.9 $\pm$ 5.6 & 60.3 $\pm$ 5.8  & \textbf{68.5 $\pm$ 4.1} \\
        diabetes        & \underline{73.1 $\pm$ 5.4} & 69.5 $\pm$ 5.0 & 70.5 $\pm$ 7.0  & \textbf{73.6 $\pm$ 4.7} \\
        breast-w        & 95.2 $\pm$ 2.6 & 94.9 $\pm$ 2.5 & \underline{95.9 $\pm$ 2.8}  & \textbf{96.5 $\pm$ 1.9} \\
        ilpd            & \underline{59.2 $\pm$ 7.7} & 57.4 $\pm$ 5.3 & 58.9 $\pm$ 6.8  & \textbf{61.9 $\pm$ 10.4} \\
        monks2          & \underline{98.0 $\pm$ 2.5} & \underline{98.0 $\pm$ 2.6} & 78.3 $\pm$ 8.1  & \textbf{99.5 $\pm$ 0.8} \\
        climate         & \underline{71.2 $\pm$ 11.0} & 68.8 $\pm$ 12.7 & 68.1 $\pm$ 10.7 & \textbf{73.1 $\pm$ 15.6}  \\
        kc2             & 68.8 $\pm$ 6.0 & 69.0 $\pm$ 7.3 & \underline{69.2 $\pm$ 7.5} & \textbf{77.6 $\pm$ 7.9} \\
        pc1             & \underline{62.8 $\pm$ 5.4} & 62.5 $\pm$ 7.9 & 61.2 $\pm$ 9.4  & \textbf{66.6 $\pm$ 7.2} \\
        kc1             & 63.7 $\pm$ 4.0 & \underline{\textbf{66.8 $\pm$ 2.5}} & 65.0 $\pm$ 3.3 & 64.9 $\pm$ 4.9  \\
        heart           & 82.2 $\pm$ 5.3 & 82.9 $\pm$ 4.8 & \underline{83.3 $\pm$ 5.2} & \textbf{84.8 $\pm$ 6.4} \\
        tictactoe       & 99.1 $\pm$ 1.3 & 97.9 $\pm$ 2.3 & \underline{\textbf{99.9 $\pm$ 0.1}} & 94.4 $\pm$ 3.1  \\
        wdbc            & 96.1 $\pm$ 3.0 & 94.5 $\pm$ 3.5 & \underline{\textbf{96.3 $\pm$ 2.4}} & 95.6 $\pm$ 3.5  \\
        churn           & 87.5 $\pm$ 2.3 & 85.4 $\pm$ 2.4 & \underline{\textbf{88.3 $\pm$ 1.9}} & 86.3 $\pm$ 2.0  \\
        pc3             & 60.2 $\pm$ 5.8 & 59.0 $\pm$ 7.4 & \underline{60.7 $\pm$ 6.5} & \textbf{62.9 $\pm$ 6.6} \\
        biodeg          & 84.8 $\pm$ 3.4 & 82.4 $\pm$ 5.7 & \underline{\textbf{85.3 $\pm$ 2.7}} & 82.7 $\pm$ 3.7  \\
        credit          & 86.5 $\pm$ 3.9 & 85.3 $\pm$ 4.2 & \underline{\textbf{86.8 $\pm$ 4.4}} & 85.7 $\pm$ 5.2  \\
        spambase        & \underline{94.9 $\pm$ 1.6} & 93.8 $\pm$ 2.5 & \underline{\textbf{94.9 $\pm$ 1.2}}  & 92.9 $\pm$ 1.3 \\
        credit-g        & 67.5 $\pm$ 3.1 & \underline{\textbf{67.6 $\pm$ 5.4}} & 66.8 $\pm$ 4.9 & 65.7 $\pm$ 5.0  \\
        friedman        & \underline{\textbf{88.1 $\pm$ 3.9}} & 87.5 $\pm$ 3.3 & 86.4 $\pm$ 8.7 & 80.0 $\pm$ 7.4  \\
        usps            & 97.3 $\pm$ 1.8 & 96.6 $\pm$ 1.6 & \underline{\textbf{98.0 $\pm$ 1.5}} & 97.5 $\pm$ 1.6  \\
        bioresponse     & \underline{\textbf{78.2 $\pm$ 2.7}} & 77.1 $\pm$ 2.0 & 77.7 $\pm$ 3.3 & 72.9 $\pm$ 2.6  \\
        speeddating     & 68.6 $\pm$ 1.4 & 68.1 $\pm$ 1.4 & \underline{\textbf{69.1 $\pm$ 1.9}} & 65.4 $\pm$ 1.9  \\
        
        \bottomrule
    \end{tabular}
    \label{tab__gbdts_performance}
\end{table*}
\begin{table*}[htb]
    \caption{Classification performance of TabPFN and \mbox{RO-FIGS} on 22 tabular datasets. We report mean $\pm$ std of the test balanced accuracy across 10 folds. The highest accuracy for each dataset across all methods is bolded. TabPFN is a competitive and fast baseline, but it is not suitable for datasets with more than 100 features or 10k samples (the last three rows). Moreover, it is a transformer-based method and thus offers very limited interpretability for its decisions. 
    }
    \centering
    \begin{tabular}{lrr}
        \toprule
        Dataset & TabPFN & RO-FIGS \\
        \midrule
            
            blood       & 63.5 $\pm$ 4.6    & \textbf{68.5 $\pm$ 4.1} \\
            diabetes    & 72.0 $\pm$ 6.5    & \textbf{73.6 $\pm$ 4.7} \\
            breast-w    & \textbf{96.7 $\pm$ 2.0}    & 96.5 $\pm$ 1.9 \\
            ilpd        & 60.5 $\pm$ 5.0    & \textbf{61.9 $\pm$ 10.4} \\
            monks2      & \textbf{99.9 $\pm$ 0.1}   & 99.5 $\pm$ 0.8 \\
            climate     & \textbf{82.8 $\pm$ 8.2}    & 73.1 $\pm$ 15.6 \\
            kc2         & 69.0 $\pm$ 7.1    & \textbf{77.6 $\pm$ 7.9} \\
            pc1         & 52.5 $\pm$ 4.9    & \textbf{66.6 $\pm$ 7.2} \\
            kc1         & 58.2 $\pm$ 3.1    & \textbf{64.9 $\pm$ 4.9} \\
            heart       & 84.5 $\pm$ 7.1    & \textbf{84.8 $\pm$ 6.4} \\
            tictactoe   & \textbf{97.0 $\pm$ 2.1}    & 94.4 $\pm$ 3.1 \\
            wdbc        & \textbf{97.7 $\pm$ 2.2}    & 95.6 $\pm$ 3.5 \\
            churn       & 83.5 $\pm$ 2.6    & \textbf{86.3 $\pm$ 2.0} \\
            pc3         & 52.3 $\pm$ 2.3    & \textbf{62.9 $\pm$ 6.6} \\
            biodeg      & \textbf{86.8 $\pm$ 4.1}    & 82.7 $\pm$ 3.7 \\
            credit      & \textbf{86.2 $\pm$ 4.3}    & 85.7 $\pm$ 5.2 \\
            spambase    & \textbf{94.7 $\pm$ 1.1}    & 92.9 $\pm$ 1.3 \\
            credit-g    & \textbf{68.5 $\pm$ 4.5}    & 65.7 $\pm$ 5.0 \\
            friedman    & 63.4 $\pm$ 7.3    & \textbf{80.0 $\pm$ 7.4} \\
            usps        & 50.0 $\pm$ 0.0    & \textbf{97.5 $\pm$ 1.6} \\
            bioresponse & 50.0 $\pm$ 0.0    & \textbf{72.9 $\pm$ 2.6} \\
            speeddating & 50.0 $\pm$ 0.0    & \textbf{65.4 $\pm$ 1.9} \\
            
        \bottomrule
    \end{tabular}
    \label{tab__tabpfn}
\end{table*}

\subsection{Ablation studies on RO-FIGS}
\label{sec:appendix-results:ablations}

\begin{figure*}[th]
    \centering
    \vspace{0.5cm}
    \includegraphics[width=0.9\textwidth]{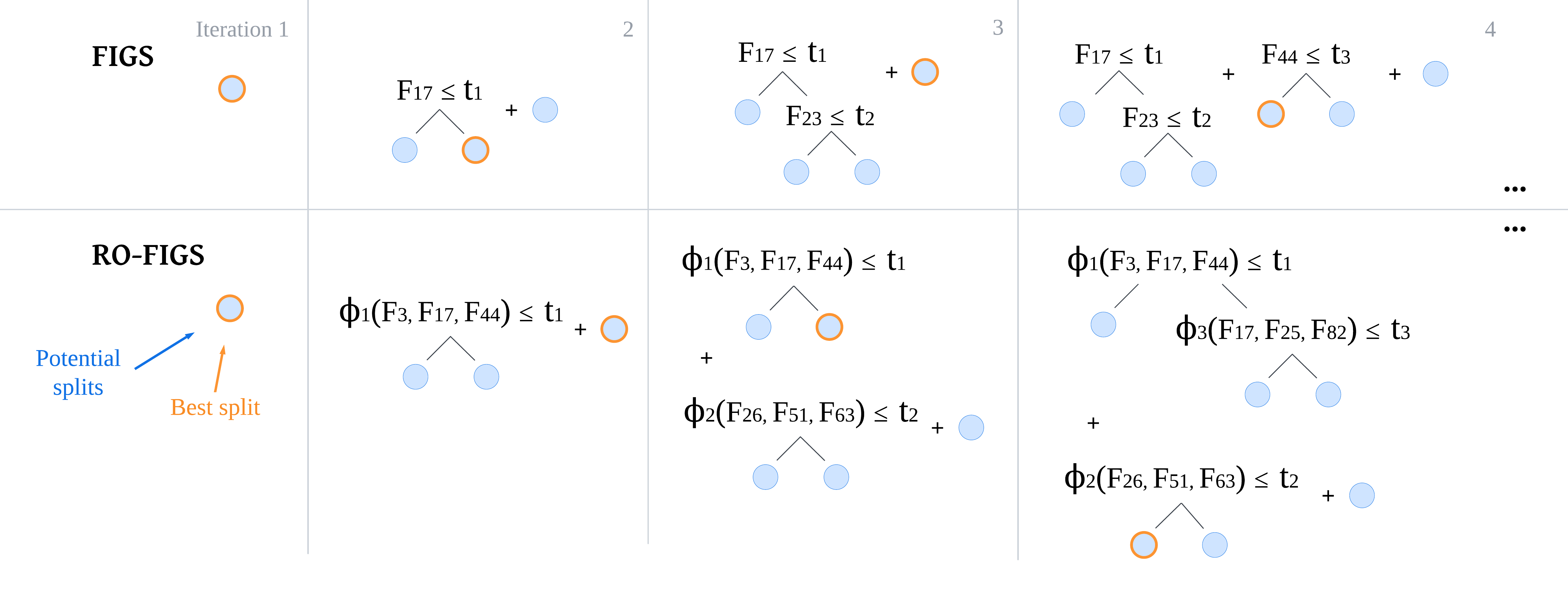}
    \hspace{1cm}
    \caption{Comparison of fitting processes of FIGS and RO-FIGS. In each iteration, both methods add one split to the model. However, \mbox{RO-FIGS} computes a linear combination of multiple randomly selected features, while FIGS builds splits with one feature. $t$, $\Phi$, $F$ denote thresholds, functions, and features, respectively. The figure has been adapted from~\cite{tan_2022_figs}.}
    \label{fig__rofigs_vs_figs} 
\end{figure*}

We first compare learning processes in FIGS and \mbox{RO-FIGS}, which extends Fig.~\ref{fig__rofigs_only}. Specifically, Fig.~\ref{fig__rofigs_vs_figs} highlights the main difference between FIGS and \mbox{RO-FIGS}: the use of univariate and oblique splits, respectively.

Table~\ref{tab__stopping_conditions} then demonstrates that utilising the total number of splits as a stopping condition (as in FIGS) leads to subpar performance. This implies that employing the minimum impurity decrease parameter (i.e., \textsc{min\_imp\_dec}) as a stopping condition for model growth is recommended over simply limiting the number of splits in the model.

\begin{figure*}[h!]
    \centering
    \includegraphics[width=0.43\textwidth]{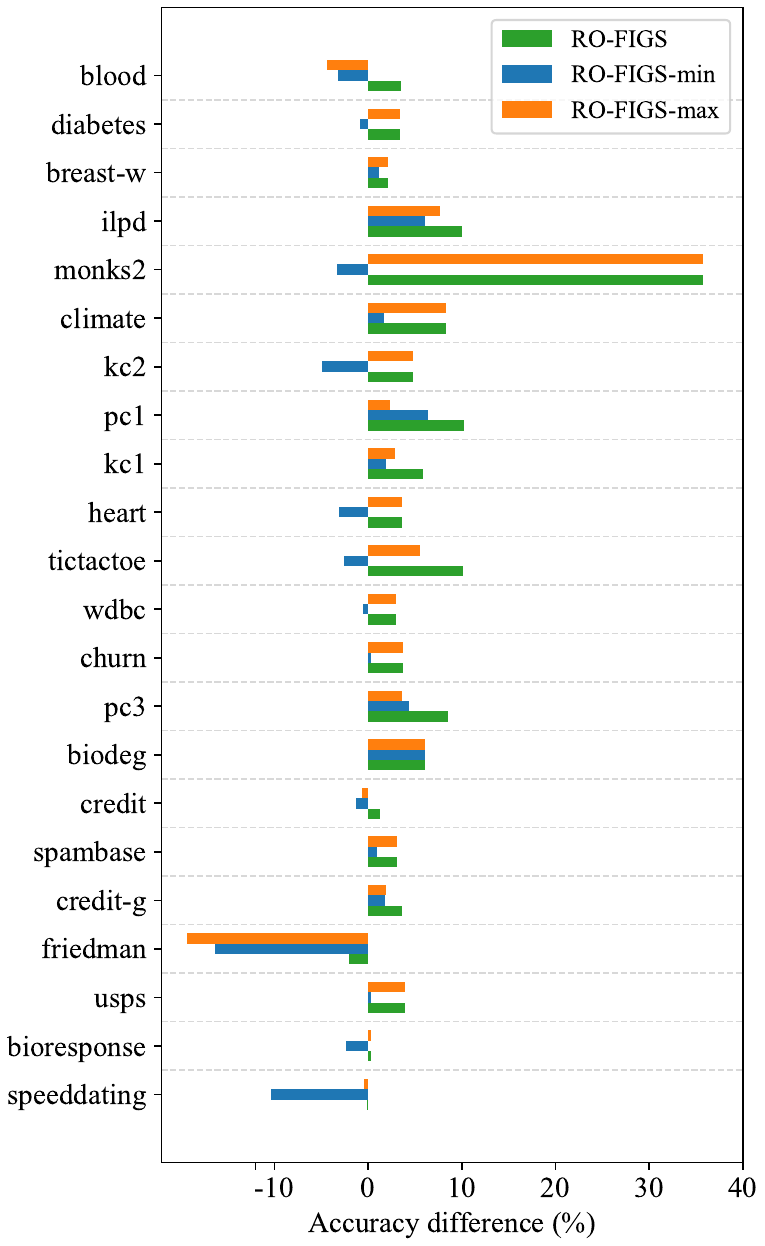}    
    \caption{The effect of oblique splits on the mean balanced accuracy across 10 folds. We use the FIGS baseline with univariate splits as a reference point (with an accuracy difference of 0\%). Each bar illustrates the difference in performance between FIGS and the corresponding method, where a positive difference indicates higher accuracy. \mbox{RO-FIGS} represents the optimal configuration of \mbox{RO-FIGS}, while \mbox{RO-FIGS-min} and \mbox{RO-FIGS-max} represent \mbox{RO-FIGS} models with \textsc{beam\_size} set to extreme values, that is, one and all. There is a clear improvement in accuracy when using oblique splits, but note that using all available features can be suboptimal.}
    \label{fig__comparing_rofigs_configurations} 
\end{figure*}

Next, we analyse how oblique splits contribute to performance. In Fig.~\ref{fig__comparing_rofigs_configurations}, FIGS, shown as a reference point at 0\%, and \mbox{RO-FIGS} correspond to values in Table~\ref{tab__performance}. 
We further investigate how the extreme values of the \textsc{beam\_size} parameter influence the performance of \mbox{RO-FIGS} models. Namely, \mbox{RO-FIGS-min} and \mbox{RO-FIGS-max} consider one or all available features at each split, respectively. 
We can see a notable increase in performance when using oblique splits (in both \mbox{RO-FIGS} and \mbox{RO-FIGS-max}), although considering all available features at each step can be suboptimal. 
Finally, we expect \mbox{RO-FIGS-min} to perform the worst as it randomly picks one feature when creating splits. We observe however that it outperforms FIGS on 11 out of 22 datasets. This suggests that using \textsc{min\_imp\_dec} as a stopping condition is sometimes better than simply limiting the maximum number of splits in the model. Note that both methods construct univariate splits.

\begin{table*}[h]
    \centering
    \caption{Performance comparison of RO-FIGS with different stopping conditions: RO-FIGS-splits uses the total number of splits (i.e., \textsc{max\_splits}, restricted to 20 like FIGS), while RO-FIGS employs the minimum impurity decrease condition (i.e., \textsc{min\_imp\_dec}). Both configurations represent the optimal setup. We report the mean $\pm$ std of the test balanced accuracy across 10 folds in the middle two columns, while the last column shows the relative increase in performance of RO-FIGS over RO-FIGS-splits. RO-FIGS, guided by the minimum impurity decrease, exhibits superior accuracy. Notably, the highest performance increase of 9.7\% is seen on the \textit{kc2} dataset.}
    \begin{tabular}{lrrr}
        \toprule
        Dataset & RO-FIGS-splits & RO-FIGS & Increase (\%) \\
        \midrule
        blood & 62.7 $\pm$ 6.4 & 68.5 $\pm$ 4.1 & 5.8 \\
        diabetes & 72.6 $\pm$ 6.4 & 73.6 $\pm$ 4.7 & 1.0 \\
        breast-w & 96.0 $\pm$ 2.0 & 96.5 $\pm$ 1.9 & 0.5 \\
        ilpd & 59.1 $\pm$ 5.0 & 61.9 $\pm$ 10.4 & 2.8 \\
        monks2 & 99.5 $\pm$ 0.8 & 99.5 $\pm$ 0.8 & 0.0 \\
        climate & 72.0 $\pm$ 15.9 & 73.1 $\pm$ 15.6 & 1.1 \\
        kc2 & 67.9 $\pm$ 6.1 & 77.6 $\pm$ 7.9 & 9.7 \\
        pc1 & 62.9 $\pm$ 5.6 & 66.6 $\pm$ 7.2 & 3.7 \\
        kc1 & 60.7 $\pm$ 4.3 & 64.9 $\pm$ 4.9 & 4.2 \\
        heart & 82.4 $\pm$ 7.6 & 84.8 $\pm$ 6.4 & 2.4 \\
        tictactoe & 91.5 $\pm$ 3.9 & 94.4 $\pm$ 3.1 & 2.9 \\
        wdbc & 95.6 $\pm$ 3.4 & 95.6 $\pm$ 3.5 & 0.0  \\
        churn & 85.5 $\pm$ 1.8 & 86.3 $\pm$ 2.0 & 0.8 \\
        pc3 & 59.1 $\pm$ 6.1 & 62.9 $\pm$ 6.6 & 3.8 \\
        biodeg & 82.6 $\pm$ 2.9 & 82.7 $\pm$ 3.7 & 0.1 \\
        credit & 85.3 $\pm$ 3.8 & 85.7 $\pm$ 5.2 & 0.4 \\
        spambase & 92.7 $\pm$ 1.5 & 92.9 $\pm$ 1.3 & 0.2 \\
        credit-g & 64.8 $\pm$ 4.2 & 65.7 $\pm$ 5.0 & 0.9 \\
        friedman & 76.0 $\pm$ 5.9 & 80.0 $\pm$ 7.4 & 4.0 \\
        usps & 97.7 $\pm$ 1.6 & 97.5 $\pm$ 1.6 & -0.2 \\
        bioresponse & 72.9 $\pm$ 2.6 & 72.9 $\pm$ 2.6 & 0.0  \\
        speeddating & 66.0 $\pm$ 1.6 & 65.4 $\pm$ 1.9 & -0.6 \\
        \bottomrule
    \end{tabular}
    \label{tab__stopping_conditions}
\end{table*}

\clearpage

\subsection{Efficiency}
\label{sec:appendix-results:efficiency}

Tables~\ref{tab__trees} and~\ref{tab__splits} show the number of trees and splits for the tree-based methods, respectively. They demonstrate that \mbox{RO-FIGS} models are compact and significantly smaller than the standard tree-based ensembles such as random forest and CatBoost. In terms of size, \mbox{RO-FIGS} models are most comparable to FIGS (limited to 20 splits) but exhibit much better performance (see Table~\ref{tab__performance}).

\begin{table*}[h!]
    \caption{The number of trees built by tree-based methods. We leave out methods that build exactly one tree and report the mean $\pm$ std of the number of trees on the test set across 10 folds. RO-FIGS builds multiple trees when the underlying data has an additive structure present, and this almost always coincides with FIGS finding the additive structure as well. The number of trees in RO-FIGS models is unsurprisingly much lower than the number of trees in ensembles and comparable to that of FIGS.}
    \centering
    \begin{tabular}{lrrrrrr}
        \toprule
        Dataset & RF & ETC & CatBoost & FIGS & EnsODT & RO-FIGS \\
        \midrule
        blood & 73.4 $\pm$ 51.2 & 135.4 $\pm$ 71.6 & 78.0 $\pm$ 78.0 & 1.0 $\pm$ 0.0 & 24.8 $\pm$ 10.4 & 1.0 $\pm$ 0.0 \\
        diabetes & 96.1 $\pm$ 70.0 & 143.4 $\pm$ 76.7 & 59.0 $\pm$ 39.8 & 1.6 $\pm$ 0.7 & 29.7 $\pm$ 17.7 & 1.0 $\pm$ 0.0 \\
        breast-w & 41.3 $\pm$ 57.7 & 47.1 $\pm$ 58.7 & 47.8 $\pm$ 46.0 & 1.0 $\pm$ 0.0 & 18.6 $\pm$ 12.9 & 1.0 $\pm$ 0.0 \\
        ilpd & 60.7 $\pm$ 58.0 & 151.5 $\pm$ 70.6 & 169.5 $\pm$ 89.0 & 1.8 $\pm$ 0.6 & 18.7 $\pm$ 11.6 & 3.0 $\pm$ 0.8 \\
        monks2 & 184.3 $\pm$ 15.2 & 176.6 $\pm$ 18.9 & 230.6 $\pm$ 19.4 & 1.0 $\pm$ 0.0 & 28.4 $\pm$ 10.9 & 2.3 $\pm$ 1.1 \\
        climate & 71.5 $\pm$ 73.5 & 9.0 $\pm$ 0.0 & 120.9 $\pm$ 57.2 & 1.1 $\pm$ 0.3 & 19.6 $\pm$ 14.7 & 2.0 $\pm$ 0.9 \\
        kc2 & 57.5 $\pm$ 78.0 & 84.0 $\pm$ 100.8 & 73.1 $\pm$ 65.5 & 1.0 $\pm$ 0.0 & 23.0 $\pm$ 8.8 & 1.0 $\pm$ 0.0 \\
        pc1 & 50.0 $\pm$ 58.4 & 109.9 $\pm$ 80.5 & 91.1 $\pm$ 75.5 & 1.6 $\pm$ 0.7 & 25.2 $\pm$ 14.5 & 2.4 $\pm$ 1.0 \\
        kc1 & 64.5 $\pm$ 66.6 & 93.1 $\pm$ 85.3 & 111.7 $\pm$ 66.3 & 1.3 $\pm$ 0.5 & 27.1 $\pm$ 17.6 & 1.4 $\pm$ 0.5 \\
        heart & 49.7 $\pm$ 71.8 & 68.4 $\pm$ 46.1 & 53.8 $\pm$ 52.3 & 3.1 $\pm$ 1.5 & 23.6 $\pm$ 13.9 & 1.0 $\pm$ 0.0 \\
        tictactoe & 173.5 $\pm$ 27.0 & 148.2 $\pm$ 60.6 & 214.0 $\pm$ 5.6 & 4.5 $\pm$ 2.7 & 41.2 $\pm$ 5.7 & 4.9 $\pm$ 1.3 \\
        wdbc & 28.5 $\pm$ 38.3 & 45.9 $\pm$ 50.1 & 68.0 $\pm$ 45.1 & 1.0 $\pm$ 0.0 & 24.1 $\pm$ 12.6 & 1.0 $\pm$ 0.0 \\
        churn & 186.6 $\pm$ 21.8 & 175.4 $\pm$ 8.2 & 92.8 $\pm$ 68.3 & 1.8 $\pm$ 0.8 & 26.8 $\pm$ 13.8 & 1.1 $\pm$ 0.3 \\
        pc3 & 94.2 $\pm$ 64.1 & 97.4 $\pm$ 93.1 & 124.0 $\pm$ 60.7 & 1.5 $\pm$ 0.5 & 14.9 $\pm$ 11.9 & 1.8 $\pm$ 0.9 \\
        biodeg & 61.1 $\pm$ 46.9 & 114.3 $\pm$ 75.6 & 105.4 $\pm$ 86.0 & 2.1 $\pm$ 0.9 & 35.0 $\pm$ 14.8 & 1.0 $\pm$ 0.0 \\
        credit & 97.3 $\pm$ 55.1 & 47.9 $\pm$ 69.5 & 49.2 $\pm$ 47.1 & 1.4 $\pm$ 0.7 & 25.0 $\pm$ 13.0 & 1.0 $\pm$ 0.0 \\
        spambase & 37.1 $\pm$ 29.6 & 162.9 $\pm$ 56.9 & 79.1 $\pm$ 44.0 & 2.5 $\pm$ 1.4 & 19.1 $\pm$ 11.1 & 2.9 $\pm$ 0.7 \\
        credit-g & 156.9 $\pm$ 60.6 & 151.1 $\pm$ 65.4 & 131.7 $\pm$ 81.1 & 1.6 $\pm$ 1.0 & 31.3 $\pm$ 16.1 & 4.6 $\pm$ 1.2 \\
        friedman & 118.6 $\pm$ 49.4 & 130.5 $\pm$ 76.3 & 143.8 $\pm$ 95.1 & 1.5 $\pm$ 0.5 & 24.9 $\pm$ 8.2 & 1.2 $\pm$ 0.4 \\
        usps & 27.7 $\pm$ 18.7 & 27.6 $\pm$ 17.1 & 70.8 $\pm$ 67.1 & 1.0 $\pm$ 0.0 & 28.0 $\pm$ 8.2 & 1.0 $\pm$ 0.0 \\
        bioresponse & 188.5 $\pm$ 33.7 & 144.9 $\pm$ 64.3 & 137.9 $\pm$ 66.7 & 2.7 $\pm$ 1.1 & 34.9 $\pm$ 11.2 & 2.0 $\pm$ 0.7 \\
        speeddating & 53.9 $\pm$ 72.8 & 143.0 $\pm$ 72.9 & 178.3 $\pm$ 71.5 & 2.6 $\pm$ 1.0 & 23.7 $\pm$ 9.6 & 2.6 $\pm$ 0.5 \\
        
        \bottomrule
    \end{tabular}
    \label{tab__trees}
\end{table*}
\begin{table*}[bt]
    \caption{The number of splits in tree-based models. We report the mean $\pm$ std of the number of splits on the test set across 10 folds. RO-FIGS has a surprisingly low number of splits on some datasets (e.g., \textit{diabetes}, \textit{kc2}, \textit{heart}) and is comparable to single decision trees and FIGS models.}
    \centering
    \resizebox{\textwidth}{!}{
    \begin{tabular}{lrrrrrrrrrr}
        \toprule
        Dataset & DT & MT & OT & ODT & RF & ETC & CatBoost & FIGS & EnsODT & RO-FIGS \\
        \midrule
        blood & 29.5 $\pm$ 18.1 & 7.5 $\pm$ 3.9 & 7.0 $\pm$ 0.0 & 12.9 $\pm$ 6.5 & 1710.2 $\pm$ 1600.9 & 11192.1 $\pm$ 7989.8 & 1082.1 $\pm$ 976.4 & 6.7 $\pm$ 5.4 & 348.1 $\pm$ 143.6 & 2.0 $\pm$ 0.0 \\
        diabetes & 31.1 $\pm$ 19.1 & 19.6 $\pm$ 5.6 & 7.0 $\pm$ 0.0 & 21.6 $\pm$ 19.0 & 4461.5 $\pm$ 3693.3 & 13844.3 $\pm$ 8236.0 & 978.0 $\pm$ 1003.8 & 7.1 $\pm$ 4.4 & 805.1 $\pm$ 697.5 & 1.0 $\pm$ 0.0 \\
        breast-w & 9.6 $\pm$ 3.9 & 6.1 $\pm$ 1.7 & 7.0 $\pm$ 0.0 & 7.4 $\pm$ 5.2 & 242.0 $\pm$ 346.4 & 746.5 $\pm$ 1181.3 & 583.2 $\pm$ 1080.4 & 8.4 $\pm$ 6.1 & 92.1 $\pm$ 79.5 & 4.6 $\pm$ 1.6 \\
        ilpd & 16.3 $\pm$ 7.8 & 17.0 $\pm$ 6.3 & 7.0 $\pm$ 0.0 & 13.9 $\pm$ 6.2 & 1356.9 $\pm$ 1726.4 & 13946.2 $\pm$ 7690.0 & 2399.6 $\pm$ 2449.8 & 11.9 $\pm$ 6.9 & 376.0 $\pm$ 284.4 & 75.0 $\pm$ 0.0 \\
        monks2 & 43.5 $\pm$ 6.4 & 12.3 $\pm$ 2.9 & 7.0 $\pm$ 0.0 & 16.9 $\pm$ 9.7 & 12997.4 $\pm$ 3299.3 & 17884.9 $\pm$ 7510.0 & 6439.9 $\pm$ 895.7 & 17.6 $\pm$ 3.1 & 428.8 $\pm$ 221.3 & 11.5 $\pm$ 5.6 \\
        climate & 10.4 $\pm$ 4.4 & 7.7 $\pm$ 1.8 & 7.0 $\pm$ 0.0 & 9.3 $\pm$ 3.8 & 1153.5 $\pm$ 1368.3 & 9.0 $\pm$ 0.0 & 970.5 $\pm$ 362.3 & 7.1 $\pm$ 3.3 & 124.3 $\pm$ 104.4 & 14.3 $\pm$ 5.3 \\
        kc2 & 9.8 $\pm$ 10.1 & 4.2 $\pm$ 2.0 & 7.0 $\pm$ 0.0 & 13.8 $\pm$ 12.5 & 505.4 $\pm$ 1029.1 & 3520.7 $\pm$ 4598.1 & 1312.7 $\pm$ 2038.7 & 4.3 $\pm$ 4.2 & 305.2 $\pm$ 172.9 & 1.0 $\pm$ 0.0 \\
        pc1 & 21.9 $\pm$ 10.0 & 3.5 $\pm$ 1.4 & 7.0 $\pm$ 0.0 & 21.0 $\pm$ 12.1 & 1352.1 $\pm$ 2298.3 & 9208.8 $\pm$ 7772.3 & 1584.7 $\pm$ 1249.0 & 9.6 $\pm$ 6.9 & 530.8 $\pm$ 356.5 & 61.7 $\pm$ 10.3 \\
        kc1 & 70.4 $\pm$ 39.5 & 13.1 $\pm$ 2.9 & 7.0 $\pm$ 0.0 & 18.3 $\pm$ 14.4 & 5047.0 $\pm$ 8846.3 & 20019.2 $\pm$ 24310.8 & 2254.7 $\pm$ 1343.0 & 12.1 $\pm$ 5.7 & 1160.6 $\pm$ 1103.8 & 75.0 $\pm$ 0.0 \\
        heart & 8.2 $\pm$ 4.6 & 9.9 $\pm$ 1.8 & 7.0 $\pm$ 0.0 & 9.9 $\pm$ 5.2 & 557.7 $\pm$ 1354.8 & 667.5 $\pm$ 455.6 & 742.7 $\pm$ 1659.3 & 7.0 $\pm$ 5.3 & 232.0 $\pm$ 188.2 & 1.0 $\pm$ 0.0 \\
        tictactoe & 32.8 $\pm$ 14.6 & 0.0 $\pm$ 0.0 & 7.0 $\pm$ 0.0 & 29.4 $\pm$ 10.0 & 17075.7 $\pm$ 7980.9 & 16977.5 $\pm$ 8817.3 & 4787.6 $\pm$ 1135.9 & 18.8 $\pm$ 1.2 & 1072.4 $\pm$ 244.4 & 74.7 $\pm$ 0.9 \\
        wdbc & 5.9 $\pm$ 2.9 & 7.4 $\pm$ 3.5 & 7.0 $\pm$ 0.0 & 4.9 $\pm$ 6.5 & 250.8 $\pm$ 309.9 & 982.7 $\pm$ 1657.8 & 650.7 $\pm$ 543.4 & 5.0 $\pm$ 3.0 & 177.5 $\pm$ 180.2 & 3.3 $\pm$ 1.6 \\
        churn & 89.9 $\pm$ 47.5 & 53.4 $\pm$ 23.6 & 7.0 $\pm$ 0.0 & 54.9 $\pm$ 31.3 & 29896.6 $\pm$ 3760.8 & 50553.5 $\pm$ 6254.8 & 1539.1 $\pm$ 680.9 & 16.5 $\pm$ 3.4 & 1939.1 $\pm$ 1195.6 & 7.6 $\pm$ 1.5 \\
        pc3 & 46.5 $\pm$ 18.4 & 22.4 $\pm$ 5.9 & 7.0 $\pm$ 0.0 & 31.9 $\pm$ 18.3 & 4160.6 $\pm$ 3864.9 & 12664.5 $\pm$ 13116.7 & 2374.8 $\pm$ 2167.2 & 12.7 $\pm$ 5.8 & 508.4 $\pm$ 488.7 & 74.6 $\pm$ 1.3 \\
        biodeg & 28.6 $\pm$ 13.9 & 22.1 $\pm$ 10.2 & 7.0 $\pm$ 0.0 & 20.5 $\pm$ 18.1 & 2963.5 $\pm$ 2593.0 & 10230.1 $\pm$ 8861.9 & 1454.1 $\pm$ 1161.7 & 12.4 $\pm$ 7.8 & 1271.8 $\pm$ 1085.7 & 1.5 $\pm$ 0.5 \\
        credit & 17.1 $\pm$ 11.9 & 17.8 $\pm$ 3.0 & 7.0 $\pm$ 0.0 & 16.1 $\pm$ 11.7 & 2166.9 $\pm$ 2690.3 & 472.6 $\pm$ 289.9 & 1446.3 $\pm$ 1623.7 & 4.8 $\pm$ 5.2 & 347.3 $\pm$ 226.8 & 1.0 $\pm$ 0.0 \\
        spambase & 69.8 $\pm$ 41.1 & 34.7 $\pm$ 12.8 & 7.0 $\pm$ 0.0 & 32.5 $\pm$ 16.6 & 3294.5 $\pm$ 1900.7 & 41196.7 $\pm$ 19747.7 & 1057.7 $\pm$ 302.1 & 16.2 $\pm$ 2.7 & 1152.7 $\pm$ 770.1 & 67.1 $\pm$ 9.6 \\
        credit-g & 46.3 $\pm$ 23.4 & 21.5 $\pm$ 2.0 & 7.0 $\pm$ 0.0 & 29.9 $\pm$ 25.4 & 14847.1 $\pm$ 10333.6 & 21764.1 $\pm$ 19358.2 & 1922.3 $\pm$ 1344.5 & 12.3 $\pm$ 4.1 & 934.2 $\pm$ 983.3 & 33.0 $\pm$ 5.7 \\
        friedman & 14.7 $\pm$ 8.7 & 3.2 $\pm$ 0.4 & 7.0 $\pm$ 0.0 & 13.8 $\pm$ 7.1 & 3213.1 $\pm$ 2491.3 & 8503.3 $\pm$ 6733.3 & 1423.5 $\pm$ 1036.9 & 8.1 $\pm$ 3.4 & 407.9 $\pm$ 204.5 & 3.5 $\pm$ 1.1 \\
        usps & 20.0 $\pm$ 7.0 & 2.5 $\pm$ 0.5 & 7.0 $\pm$ 0.0 & 3.3 $\pm$ 2.7 & 563.5 $\pm$ 343.2 & 1635.4 $\pm$ 1455.6 & 651.2 $\pm$ 480.3 & 12.8 $\pm$ 6.0 & 280.6 $\pm$ 277.6 & 2.4 $\pm$ 0.8 \\
        bioresponse & 52.9 $\pm$ 18.4 & 63.7 $\pm$ 7.7 & 7.0 $\pm$ 0.0 & 74.1 $\pm$ 84.7 & 37456.3 $\pm$ 11098.1 & 40008.3 $\pm$ 25743.0 & 2113.2 $\pm$ 1808.4 & 12.0 $\pm$ 5.2 & 3145.8 $\pm$ 2438.1 & 5.1 $\pm$ 2.0 \\
        speeddating & 101.5 $\pm$ 67.6 & 38.0 $\pm$ 4.0 & 7.0 $\pm$ 0.0 & 67.3 $\pm$ 42.0 & 17446.1 $\pm$ 26780.3 & 103295.8 $\pm$ 64515.4 & 4255.5 $\pm$ 2865.4 & 7.2 $\pm$ 4.6 & 2187.7 $\pm$ 1208.7 & 13.1 $\pm$ 2.8 \\

        \bottomrule
    \end{tabular}
    }
    \label{tab__splits}
\end{table*}

\clearpage

\subsection{Expressiveness}
\label{sec:appendix-results:expressiveness}

Here, we further discuss the expressiveness of the \mbox{RO-FIGS} models. Due to the robustness of tree-based models to uninformative features, feature importance information is often extracted from them. To extract such information from \mbox{RO-FIGS} models, we first count combinations of features that appear together in the splits. We then analyse this jointly with the SHAP summary plot, which illustrates how each feature contributed to the model. Next, we analyse SHAP plots of baseline methods. In particular, we choose FIGS due to its similarity to \mbox{RO-FIGS} (w.r.t.\ the framework and size of the models), and CatBoost as the best-performing baseline.

\begin{figure}[htb!]
    \centering
    \begin{minipage}[b]{0.5\textwidth}
        \centering
        \includegraphics[width=\textwidth]{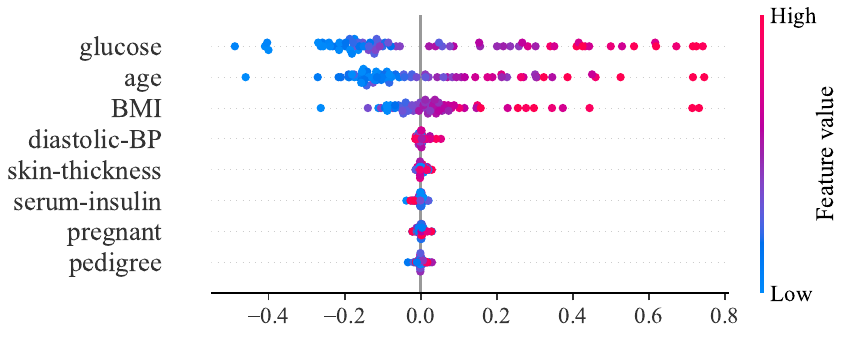}
        \vspace{-0.3cm}
        \centerline{(a) RO-FIGS}
        \label{diabA}
    \end{minipage}
    
    \vspace{0.5cm}
    
    \begin{minipage}[b]{0.5\textwidth}
        \centering
        \includegraphics[width=\textwidth]{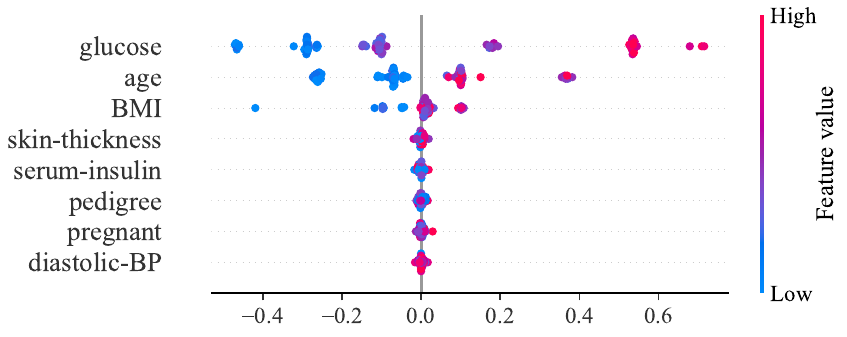}
        \vspace{-0.3cm}
        \centerline{(b) FIGS}
        \label{diabB}
    \end{minipage}
    
    \vspace{0.5cm}
    
    \begin{minipage}[b]{0.5\textwidth}
        \centering
        \includegraphics[width=\textwidth]{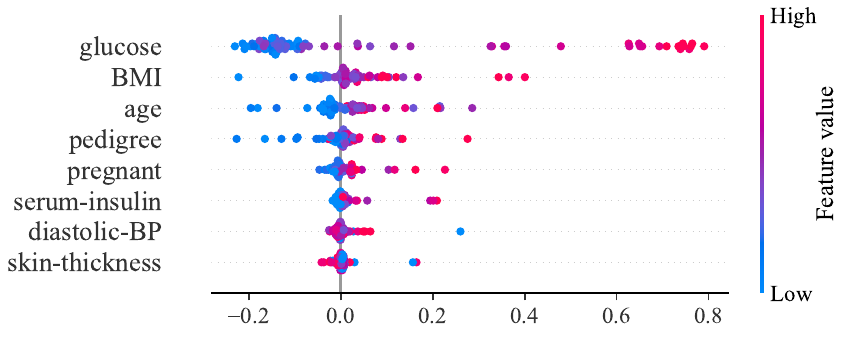}
        \vspace{-0.3cm}
        \centerline{(c) CatBoost}
        \label{diabC}
    \end{minipage}
    
    \caption{SHAP summary plots of different models on the first fold of the \textit{diabetes} dataset all implicate that features \textit{glucose}, \textit{age}, and \textit{BMI} contribute to their predictions the most. \mbox{RO-FIGS} with the accuracy of 73.6\% (see Table~\ref{tab__performance}) outperforms all baselines on this dataset with only one split, demonstrating that a linear combination of features \textit{glucose}, \textit{age}, and \textit{BMI} is optimal.}
    \label{fig__appxc_diabetes}
\end{figure}

\begin{figure}[tb!]
    \centering
    \begin{minipage}[b]{0.5\textwidth}
        \centering
        \includegraphics[width=\textwidth]{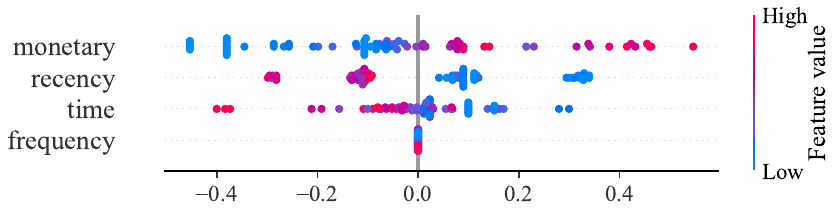}
        \vspace{-0.3cm}
        \centerline{(a) RO-FIGS}
    \end{minipage}

    \vspace{0.5cm}
    
    \begin{minipage}[b]{0.5\textwidth}
        \centering
        \includegraphics[width=\textwidth]{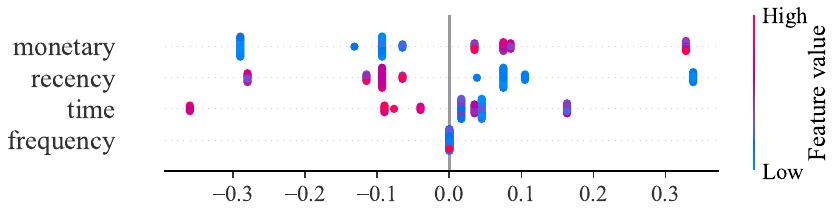}
        \vspace{-0.3cm}
        \centerline{(b) FIGS}
    \end{minipage}
    
    \vspace{0.5cm}
    
    \begin{minipage}[b]{0.5\textwidth}
        \centering
        \includegraphics[width=\textwidth]{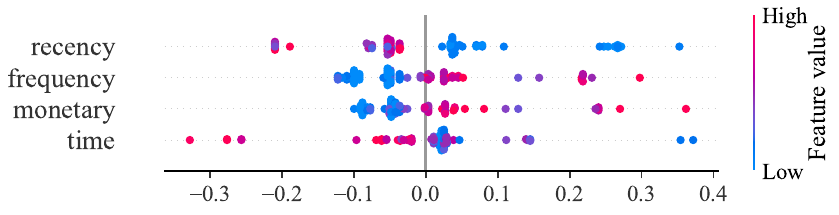}
        \vspace{-0.3cm}
        \centerline{(c) CatBoost}
    \end{minipage}
    
    \caption{The same as Fig.~\ref{fig__appxc_diabetes} but on the \textit{blood} dataset. Three methods generally disagree on the importance of features: neither of the FIGS-based models split on \textit{frequency}, but this feature is the second most important for CatBoost. 
    With only 2 splits, one using \textit{recency} and the other a linear combination of \textit{monetary} and \textit{time}), \mbox{RO-FIGS} is able to highlight combinations of features. It also outperforms all baselines, suggesting that this is a highly informative combination of features.}
    \label{fig__appxc_blood}
\end{figure}

Figs.~\ref{fig__appxc_diabetes} and \ref{fig__appxc_blood} present SHAP summary plots for \mbox{RO-FIGS}, FIGS, and CatBoost models on two datasets, \textit{diabetes} and \textit{blood}, respectively. Note that these plots are computed on fold 0, but similar behaviour is observed across all folds. Shap values are estimated using Kernel SHAP, where the nsamples parameter is set to 100. Furthermore, we utilise \texttt{shap.sample} to generate a background summary with 100 samples. 

We observe that the oblique nature of splits in \mbox{RO-FIGS} reflect feature interactions, which makes them more expressive than models with univariate trees. Furthermore, such insights cannot be learnt from feature importance tools such as SHAP. For example, on the \textit{diabetes} dataset, \mbox{RO-FIGS} builds models with only one split involving a linear combination of three features. Similarly, on the \textit{blood} dataset, \mbox{RO-FIGS} model consists of two splits, one of which is a linear combination of two features \textit{monetary} and \textit{time}.

\end{appendices}

\end{document}